\documentclass[letterpaper]{article}
\usepackage{aaai2027}
\usepackage[hyphens]{url}
\usepackage{natbib}
\usepackage[utf8]{inputenc}
\usepackage{caption}
\usepackage{graphicx}
\usepackage{amsmath}
\usepackage{amsthm}
\usepackage{booktabs}
\usepackage{algorithm}
\usepackage{algorithmic}
\usepackage{amssymb}
\usepackage{multirow}
\usepackage{array}
\usepackage{placeins}

\newcolumntype{L}[1]{>{\raggedright\arraybackslash}p{#1}}
\newcolumntype{C}[1]{>{\centering\arraybackslash}p{#1}}
\newcolumntype{R}[1]{>{\raggedleft\arraybackslash}p{#1}}

\title{From ``What-If'' to ``What-Is'': Counterfactual Thinking-Inspired Semantic Alignment for Visual Brain Decoding}

\author{
Kaitao Yan\textsuperscript{\rm 1},
Chi Liu\textsuperscript{\rm 1},
Congcong Zhu\textsuperscript{\rm 1},
Huajie Chen\textsuperscript{\rm 1},\\
Gengshen Wu\textsuperscript{\rm 1},
Minghao Wang\textsuperscript{\rm 1},
Xiaotong Han\textsuperscript{\rm 2},
Tianqing Zhu\textsuperscript{\rm 1}
}

\affiliations{
\textsuperscript{\rm 1}Faculty of Data Science, City University of Macau\\
\textsuperscript{\rm 2}Zhongshan Ophthalmic Center, Sun Yat-sen University
}

\begin{document}
\maketitle

\begin{abstract}
Visual brain decoding aims to reconstruct the visual content
perceived by a person from neural measurements such as fMRI,
providing a computational approach to studying how visual
information is represented in the brain. Recent multimodal
representations and diffusion priors have improved
reconstruction realism. However, a visually plausible
reconstruction may still contain incorrect objects,
attributes, or relations because a strong generative prior can
complete content that is not sufficiently specified by the
decoded representation. Moreover, conventional reconstruction
metrics mainly assess the final image and may therefore
obscure such semantic errors.
We propose \emph{ConceptAlign}, a counterfactual semantic
alignment framework for visual brain decoding. ConceptAlign
pools decoded visual tokens and projects them into a frozen
text-embedding space, where the representation is aligned with
the ground-truth caption and separated from scene-preserving
near-miss alternatives. Generated offline by an LLM, these
alternatives modify one critical object, attribute, or
relation while retaining the scene. A margin-based objective
then learns fine-grained semantic boundaries between the
observed stimulus and plausible but incorrect interpretations,
without requiring LLM calls during inference. We introduce a
systematic three-level semantic evaluation framework covering
foundational discriminability, counterfactual description
discrimination, and representational geometry.
Experiments on the Natural Scenes Dataset show that
ConceptAlign improves multiple reconstruction measures,
counterfactual semantic discrimination, and representational
alignment over the MindEye2 backbone. Matched negative-source
ablations, independent LLM and human-written alternatives,
and human evaluation support the effectiveness and robustness
of the supervision, with favorable patterns in fine-grained
conflicts, limited-data decoding, and cross-subject structure.
\end{abstract}
\section{Introduction}
\begin{figure}[!t]
    \centering
    \includegraphics[width=\columnwidth]
    {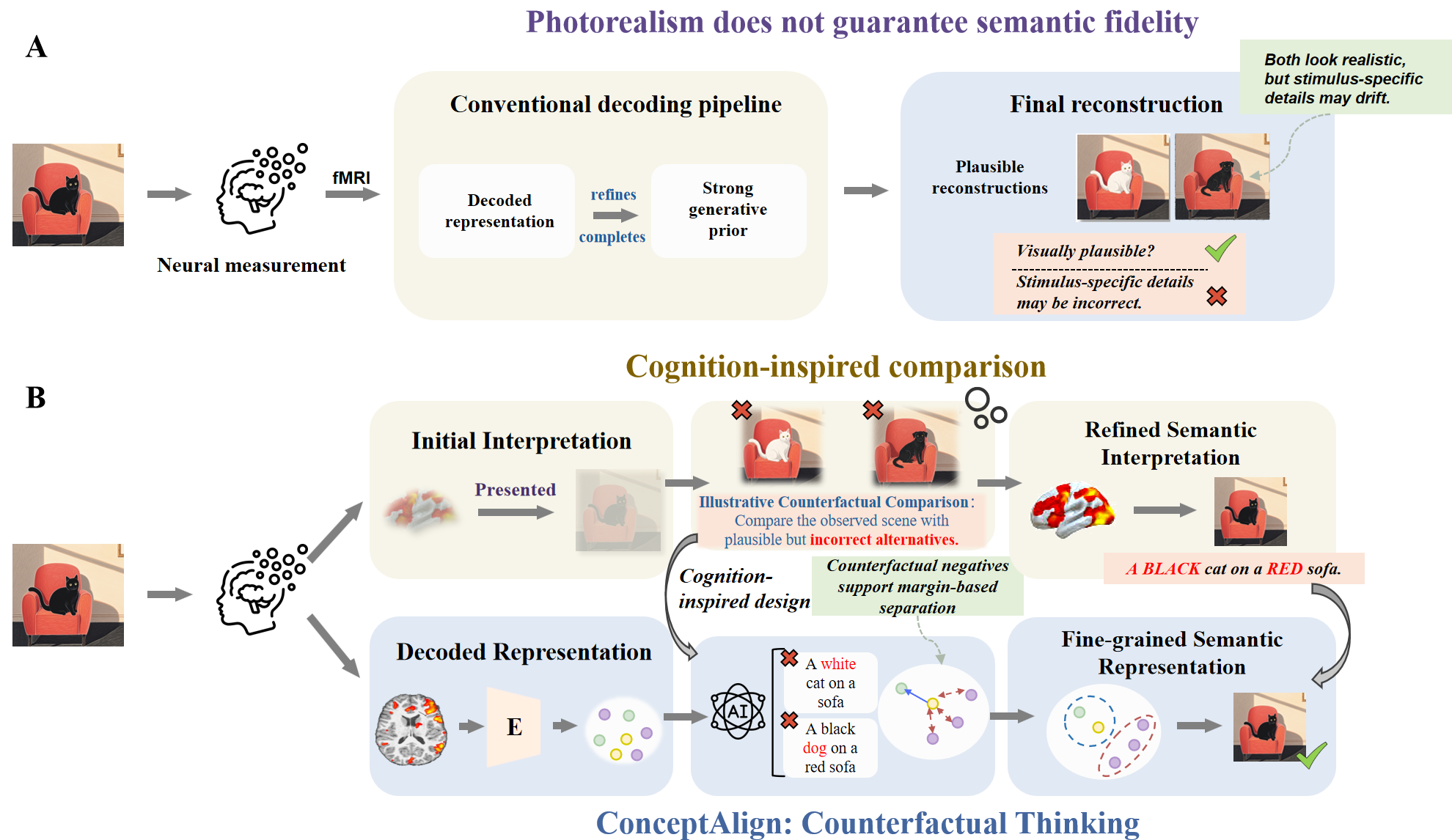}

    \caption{
    \textbf{Motivation and overview of ConceptAlign.}
    (\textbf{A}) Conventional visual brain decoding can
    produce photorealistic reconstructions even when
    stimulus-specific semantic details drift.
    (\textbf{B}) Inspired by counterfactual thinking,
    ConceptAlign distinguishes the observed scene from
    plausible but incorrect alternatives to learn
    semantic boundaries.
    }
    \label{fig:motivation}
\end{figure}

Non-invasive visual brain decoding aims to recover the visual
content perceived by a person from neural measurements such
as fMRI, providing a computational window into how visual
information is represented in the brain
~\cite{chen2023seeing,wang2024mindbridge}. Recent advances in
diffusion models and multimodal representations such as CLIP
have substantially improved the visual quality and semantic
coherence of reconstructed images
~\cite{wei2025more,liu2024eeg2video}. Representative methods,
including MindEye2~\cite{scotti2024mindeye2},
MindBridge~\cite{wang2024mindbridge}, and
NeuroPictor~\cite{huo2024neuropictor}, map fMRI responses into
pretrained visual representation spaces and use generative
priors to reconstruct the viewed scenes. As a result,
decoders can produce increasingly realistic and visually
faithful reconstructions. However, a visually plausible result
may still contain an incorrect object, attribute, relation,
count, or spatial detail~\cite{miliotou2023generative}. Such
errors are easy to overlook when a strong generative model
produces a coherent and natural-looking scene. Recent analysis
also suggests that text-guided reconstruction can combine
category-level prediction with diffusion-based completion,
causing the generated image to drift from the
stimulus~\cite{shirakawa2025spurious}. As illustrated in
Figure~\ref{fig:motivation}(A), a strong generative prior may
produce a coherent and photorealistic reconstruction while
changing stimulus-specific semantic details. This gap
motivates a closer examination of whether the decoded
representation preserves the semantic facts of the observed
scene, rather than only whether the final image appears
realistic.

Research in developmental psychology and cognitive science
suggests that people can interpret a scene by comparing it
with plausible alternatives and rejecting those that do not
fit the evidence, a process commonly referred to as
\emph{counterfactual thinking}
~\cite{weisberg2013pretense,sobel2014development}. This idea
is related to the distinction between actual
\textit{what-is} and hypothetical \textit{what-if}
situations~\cite{kominsky2021trajectory,
nyhout2019development}. We use this comparison as a computational design principle
for visual brain decoding: the decoded representation should
match the observed scene and reject a nearby description that
changes one important fact. Figure~\ref{fig:motivation}(B)
summarizes how this comparison is translated into targeted
semantic supervision.

The comparison between an observed scene and its plausible
alternatives can be framed as a contrastive discrimination
problem. However, contrastive learning often uses unrelated
negatives, which may be too easy to teach fine-grained
differences in objects, attributes, and
relations~\cite{robinson2021contrastive}. Since LLMs can
generate plausible alternatives
~\cite{webb2024evidence,palta2025investigating}, we use one
offline to create captions that retain the scene while
changing an important visual fact. These near-miss captions
translate the counterfactual comparison into targeted
training signals, requiring the decoder to distinguish what
was observed from a plausible but incorrect alternative.

\emph{ConceptAlign} projects decoded visual tokens into a
frozen text space and learns to separate the correct
description from scene-preserving near-miss alternatives.
ConceptAlign contributes a counterfactual semantic objective
for fMRI decoding, together with a matched four-way
comparison of negative sources. Because conventional
reconstruction metrics may obscure semantic errors introduced
or corrected by generative priors, we further propose a
systematic three-level evaluation framework covering
foundational discriminability, counterfactual description
discrimination, and representational geometry through 2AFC,
CCD-hard, and RSA.

Experiments on the Natural Scenes Dataset show that
ConceptAlign improves several reconstruction measures over
MindEye2 and achieves higher CCD-H and representational
alignment in the four-subject analysis. Further results show
favorable patterns in fine-grained semantic conflicts,
limited-data decoding, and cross-subject structure. Matched
ablations, Cross-LLM testing, human-written alternatives, and
human audits provide evidence for the effectiveness and
robustness of scene-preserving counterfactual supervision.

\section{Related Work}

\subsection{Visual Brain Decoding}

Visual brain decoding has moved from linear feature mapping
to deep priors and multimodal alignment
~\cite{wang2024mindbridge}. Recent fMRI methods combine
subject-specific mappings, CLIP-like representations, and
diffusion priors to recover both semantic and visual
information~\cite{chen2023seeing,scotti2024mindeye2,
wang2024mindbridge,huo2024neuropictor}. Mapping brain
activity into a strong pretrained representation greatly
improves object recognition and visual coherence, while
diffusion models provide realistic textures and scene
completion. Most training objectives, however, focus on
matching the decoded feature with its paired target. They do
not explicitly require the decoder to distinguish that target
from a nearby description that changes one important fact.

This distinction matters because a generative prior can
produce a scene when the decoded representation does not
specify its details. Shirakawa
et al.~\cite{shirakawa2025spurious} showed that realistic
text-guided reconstructions may combine category prediction
with generative hallucination. Their analysis highlights the
need to examine semantic agreement with the stimulus
alongside reconstruction quality.

A separate line of fMRI research utilizes negative examples
for post-hoc model explanation. For instance, DreaMR modifies
fMRI inputs to explain predictions made by deep
classifiers~\cite{bedel2024dreamr}, but this is fundamentally
distinct from our work in both motivation and
function---ConceptAlign is inspired by human cognitive
reasoning mechanisms, using plausible, scene-preserving
semantic alternatives as training supervision for visual
decoding.

\subsection{Contrastive Learning for Concept Alignment}

Contrastive learning brings matched representations closer
and separates unmatched ones, so selecting negatives is
important. Unrelated in-batch negatives are easy to
distinguish and may not teach the small differences needed
for fine-grained recognition~\cite{wu2022rethinking,
robinson2021contrastive,kalantidis2020hard}. Hard-negative
methods select more similar examples, but visual or embedding
similarity alone does not ensure a visually grounded
semantic conflict~\cite{yuksekgonul2023bags,
thrush2022winoground}.

Synthetic negative captions have been widely explored to
improve compositional learning in general vision--language
models. TripletCLIP, for example, trains CLIP with generated
vision--language negatives to strengthen compositional
discrimination~\cite{patel2024tripletclip}. These approaches
primarily use synthetic captions as training augmentations
for general-purpose image--text representation learning. They
are not designed to model the human process of comparing an
observed scene with a plausible but incorrect interpretation.
ConceptAlign instead introduces this comparison into visual
brain decoding, using targeted negative captions to supervise
representations decoded from fMRI.

The semantic content of a negative determines the distinction
that the representation is encouraged to learn. A random
caption can often be rejected from coarse scene information,
whereas a CLIP-nearest caption is selected by embedding
proximity and may not contain an explicit, visually grounded
conflict. Inspired by human counterfactual thinking, our
negatives preserve the scene while changing one object,
attribute, or relation. This design concentrates the training
signal on a plausible but incorrect interpretation that
differs from the observed scene in one visual fact.
\begin{figure*}[htbp!]
    \centering
    \includegraphics[width=0.95\linewidth]
    {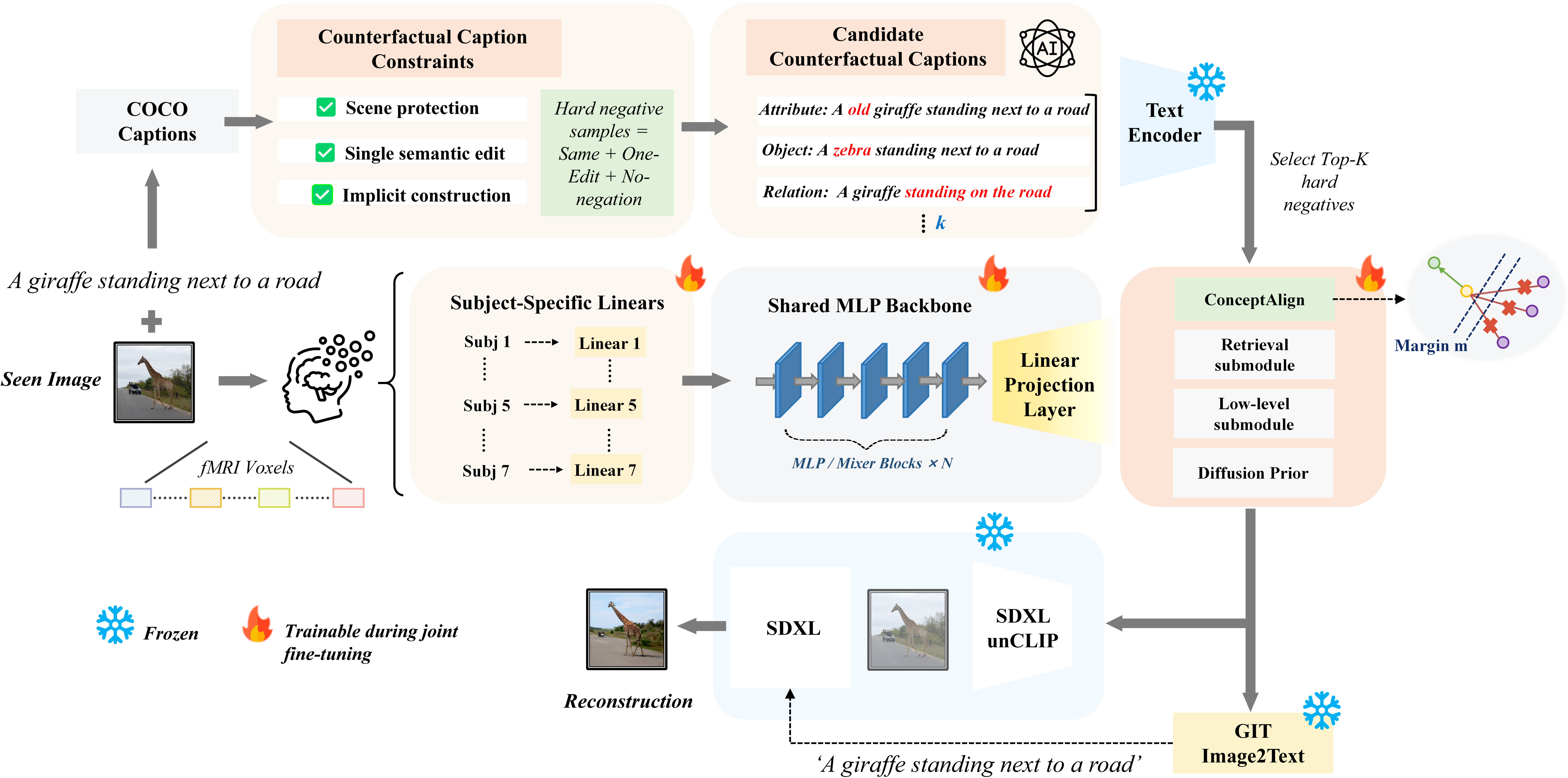}
    \caption{
ConceptAlign within the MindEye2 backbone.
Subject-specific ridge mappings and a shared decoder map fMRI
responses to visual tokens $\mathbf{Z}$. ConceptAlign pools
and projects $\mathbf{Z}$ into a frozen text space,
contrasting the ground-truth caption with counterfactual
captions generated offline by an LLM. Candidates are filtered
before Top-$K$ selection and margin-ranking supervision. The
retrieval, low-level, and diffusion-prior branches retain
their roles, and no LLM is used at inference.
}
    \label{fig:pipeline}
\end{figure*}
\section{ConceptAlign}
\label{sec:method}
ConceptAlign is inspired by human counterfactual thinking, a
cognitive mechanism where an observed scene is understood by
distinguishing it from plausible alternatives. Following this
principle, we formulate visual brain decoding as
reconstruction and fine-grained semantic discrimination. The
decoder maps fMRI responses into visual tokens for retrieval
and generation. ConceptAlign projects the pooled tokens into
a frozen text space, aligning them with the correct caption
while separating scene-preserving alternatives that change
one object, attribute, or relation.
Figure~\ref{fig:pipeline} shows our framework: retrieval,
diffusion-prior, and low-level pathways preserve
reconstruction capability, while the counterfactual alignment
pathway learns semantic boundaries between the observed scene
and plausible but incorrect interpretations.
\subsection{Visual Reconstruction Backbone}
Following MindEye2, the backbone transforms noisy,
subject-specific fMRI responses into shared visual tokens
through two stages: subject-specific functional alignment and
hierarchical visual token generation.

\paragraph{Subject-Specific Functional Alignment.}
Subject-specific ridge regression projects voxel response
$\mathbf{V}_s$ into a common space:
\begin{equation}
    \mathbf{M} = \operatorname{Ridge}_s(\mathbf{V}_s).
\end{equation}
A shared decoder maps $\mathbf{M}$ to a sequence of
visual tokens $\mathbf{Z}$.

\paragraph{Hierarchical Visual Token Generation.}
The visual tokens are trained to retain semantic and
reconstruction information. A diffusion prior aligns
$\mathbf{Z}$ with the target image representation through an
MSE objective~\cite{ramesh:hierarchical}; Bi-directional
MixCo and SoftCLIP provide contrastive supervision for
semantic retrieval~\cite{kim:mixco}; and a convolutional
upsampler maps the tokens toward the Stable Diffusion VAE
latent space for low-level reconstruction~\cite{scotti:mindeye}.
ConceptAlign is added to this backbone as an extra semantic
branch and does not replace these objectives.

\subsection{Counterfactual Thinking Inspired Learning}

ConceptAlign uses plausible but false descriptions as
near-miss alternatives. Each alternative preserves the scene
while changing one object, attribute, or relation. The module
projects visual tokens into the text space and learns a
boundary between the correct description and these
alternatives.

\subsubsection{Visual-to-Text Projection}

ConceptAlign first pools the decoded visual tokens into a
single representation and maps it into the frozen text space:
\begin{equation}
    \hat{\mathbf{t}}
    =
    \mathcal{P}_{text}
    \bigl(\operatorname{Pool}(\mathbf{Z})\bigr),
    \label{eq:text_projection}
\end{equation}
where $\operatorname{Pool}$ denotes token aggregation and
$\mathcal{P}_{text}$ is a lightweight learnable projector.
Its input follows the visual-token dimension and its output
matches the frozen text-embedding space. The projected
representation $\hat{\mathbf{t}}$ can therefore be compared
directly with the positive caption embedding
$\mathbf{t}_{pos}$ and the counterfactual embeddings
$\mathbf{t}^{-}$.

\subsubsection{Counterfactual Hard Negative Mining}

Unrelated negatives are easy to separate and provide limited
supervision for fine-grained semantic boundaries. For each
image, we start from its canonical COCO caption
$C_{gt}$~\cite{lin:coco}. An LLM generates candidates that
preserve the main scene while changing one object, attribute,
or relation. For example, ``A red apple on a wooden table''
can become ``A green apple on a wooden table'' or ``A red
ball on a wooden table.''

The prompt requires one targeted edit and no explicit
negation. Malformed, duplicate, or explicitly negated outputs
are removed before text encoding. Candidates outside the
fixed similarity window are discarded, and the top-$K$
eligible captions are selected by cosine similarity to
$\mathbf{t}_{pos}$. Similarity is only a construction filter; proximity alone
does not define a valid counterfactual. The complete
construction algorithm and exact prompt template are provided
in Appendix~\ref{supp:impl}.

Training and Cross-LLM captions use different generators.
The negative-source analysis compares positive-only,
random-caption, CLIP-nearest, and counterfactual near-miss
supervision under the same scaffold.

\begin{figure}[ht]
  \centering
  \includegraphics[width=0.95\linewidth]
  {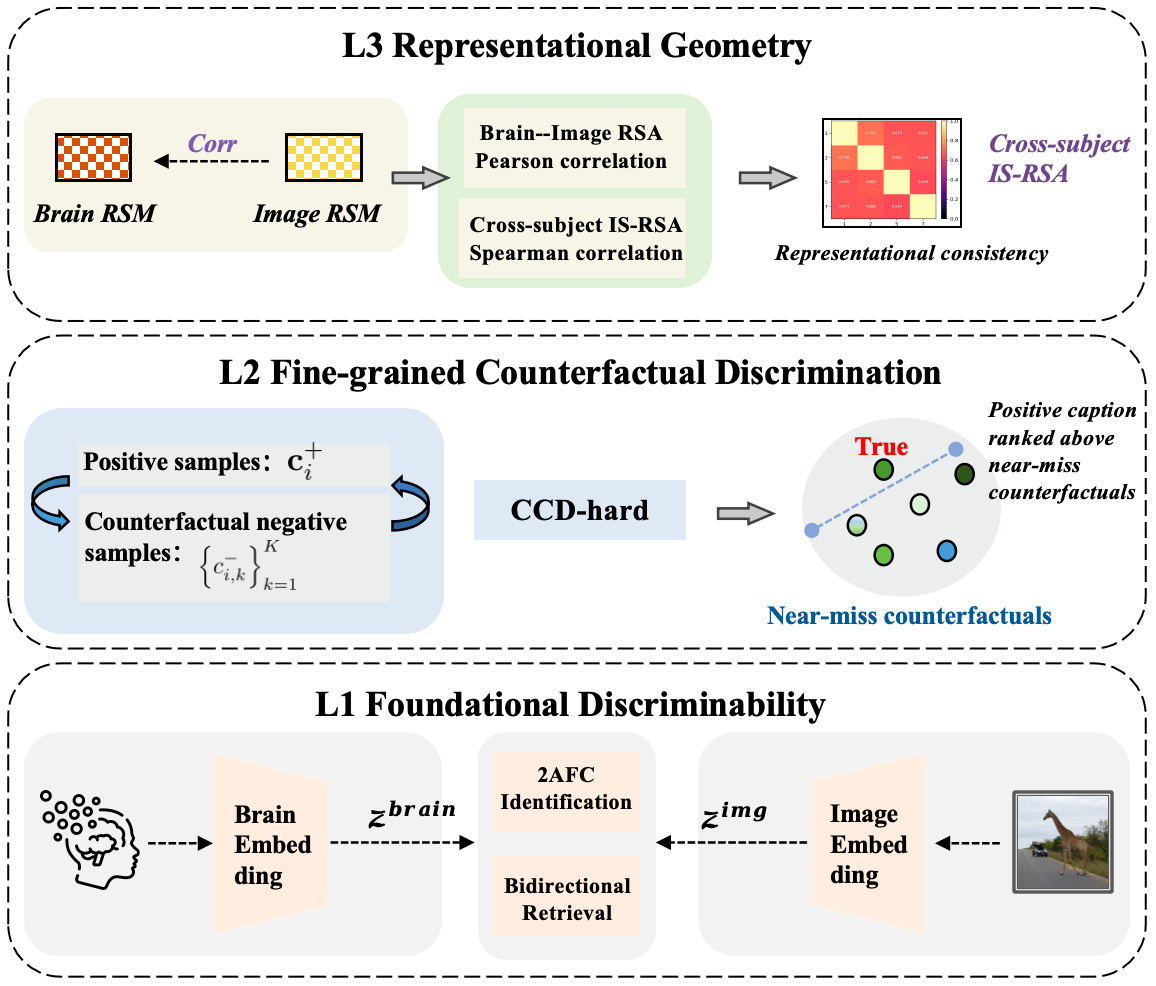}
  \caption{Semantic evaluation framework: (L1) foundational
  discriminability with retrieval and 2AFC, (L2)
  counterfactual description discrimination with CCD-hard,
  and (L3) representational geometry with RSA.}
  \label{fig:eval_framework}
\end{figure}

\subsection{Hybrid Alignment Objective}

ConceptAlign combines positive semantic alignment with
targeted rejection of counterfactual alternatives.

\paragraph{Global Semantic Alignment.}
An in-batch contrastive objective
$\mathcal{L}_{align}$ brings the predicted representation
$\hat{\mathbf{t}}_i$ closer to its positive caption embedding
$\mathbf{t}^{+}_i$ relative to the other captions in the
mini-batch. This objective aligns the decoded brain representation with
the observed image description.

\paragraph{Counterfactual Boundary Learning.}
To separate the positive caption from a near-miss negative,
we use a margin-ranking loss~\cite{ha:deep}:
\begin{equation}
\resizebox{0.84\columnwidth}{!}{$\displaystyle
\mathcal{L}_{reason}
=
\frac{1}{|B|}
\sum_{i=1}^{|B|}
\max\!\left(
0,\,
m
-
\operatorname{sim}(\hat{\mathbf{t}}_i,\mathbf{t}^{+}_i)
+
\operatorname{sim}(\hat{\mathbf{t}}_i,\mathbf{t}^{-}_i)
\right)
$}
\label{eq:reason_loss}
\end{equation}
where $B$ is the mini-batch, $m$ is the ranking margin, and
$\operatorname{sim}$ is cosine similarity between
$\ell_2$-normalized embeddings. The loss is active whenever
the positive caption is not separated from its
counterfactual negative by at least $m$.

\paragraph{Model Training.}
The backbone is first pre-trained without ConceptAlign using
the original MindEye2 objectives:
\begin{equation}
    \mathcal{L}_{pretrain}
    =
    \mathcal{L}_{prior}
    +
    \alpha_1 \mathcal{L}_{BiMixCo}
    +
    \alpha_2 \mathcal{L}_{low},
    \label{eq:pretrain_loss}
\end{equation}
where $\mathcal{L}_{prior}$ trains the diffusion prior,
$\mathcal{L}_{BiMixCo}$ supports retrieval, and
$\mathcal{L}_{low}$ supervises the low-level reconstruction
branch. The coefficients $\alpha_1$ and $\alpha_2$ balance
inherited losses.

During fine-tuning, we first keep the reconstruction branches
fixed and train ConceptAlign to establish the mapping between
decoded visual tokens and the text space. We then introduce
the counterfactual negatives and jointly optimize the
trainable components with
\begin{equation}
\begin{split}
    \mathcal{L}_{total}
    =\;&
    \lambda_{clip}\mathcal{L}_{clip}
    +
    \lambda_{prior}\mathcal{L}_{prior}
    +
    \lambda_{blur}\mathcal{L}_{blur}
    \\
    &+
    \lambda_{text}
    \left(
    \mathcal{L}_{align}
    +
    \gamma\mathcal{L}_{reason}
    \right).
\end{split}
\label{eq:total_loss}
\end{equation}
Here, $\mathcal{L}_{clip}$, $\mathcal{L}_{prior}$, and
$\mathcal{L}_{blur}$ are the inherited retrieval, diffusion
prior, and low-level reconstruction losses.
$\lambda_{clip}$, $\lambda_{prior}$,
$\lambda_{blur}$, and $\lambda_{text}$ balance the four
training branches, while $\gamma$ controls the contribution
of counterfactual ranking within the text-alignment branch.

\paragraph{Inference.}
The LLM and counterfactual captions are used to construct
supervision before and during training and are not required at
inference. The trained decoder follows the MindEye2/SDXL
reconstruction path: the diffusion prior predicts the target
image representation, the generative model produces an
initial reconstruction, and an image-to-image stage refines
the output. ConceptAlign therefore introduces no LLM calls or
caption processing during inference.

\begin{figure*}[t]
    \centering
    \includegraphics[width=\textwidth]
    {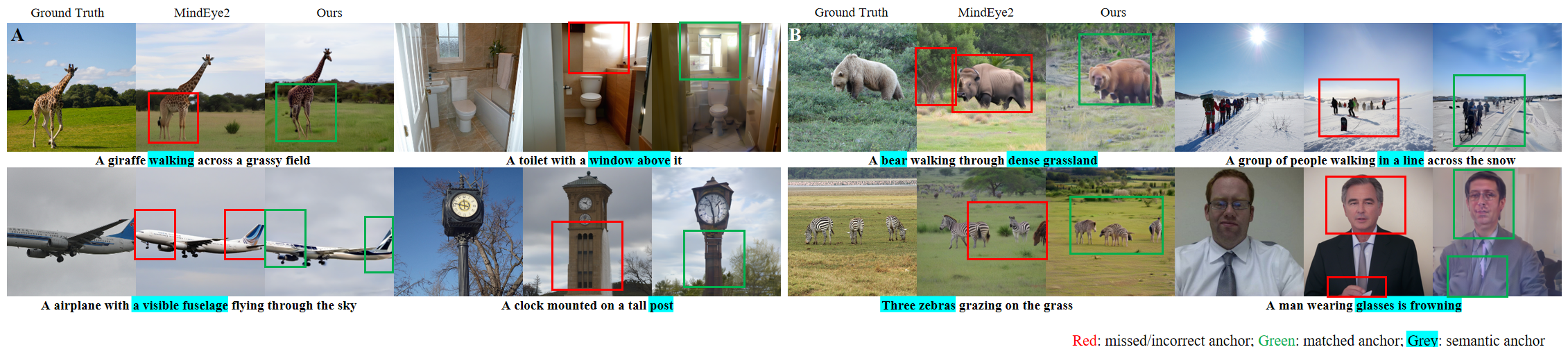}
    \caption{
\textbf{Semantic fidelity beyond visual realism.}
Each triplet shows the ground-truth image, the MindEye2
reconstruction, and the ConceptAlign reconstruction. Yellow
text marks a semantic anchor derived from the stimulus; red
boxes mark missing or incorrect anchor content, and green
boxes mark a closer match. Group A focuses on fine-grained
details when both methods recover the main scene. Group B
shows differences in object identity, count, arrangement, or
visible attributes. These selected cases illustrate semantic
differences under the stated anchors and are not used to
estimate their frequency over the full test set.
}
    \label{fig:semantic_fidelity}
\end{figure*}

\subsection{Semantic Evaluation Framework}

Conventional evaluation mainly measures the quality of final
reconstructions, whose appearance is jointly determined by
the decoded brain representation and a strong generative
prior. Consequently, visually realistic outputs may conceal
semantic errors in the underlying decoded representation. To
address this limitation, we propose the systematic three-level
semantic evaluation framework shown in
Figure~\ref{fig:eval_framework}, which examines foundational
discriminability, counterfactual description discrimination,
and representational geometry before image generation. All
vectors are $\ell_2$-normalized. Let $\mathbf{b}_i$ denote
the decoded brain representation and $\mathbf{g}_i$ the
corresponding target-image representation.

\paragraph{L1: Foundational Discriminability.}
We first measure coarse semantic matching with bidirectional
retrieval and two-way identification (2AFC).

For retrieval, the target representation $\mathbf{g}_i$ is
ranked among all $N$ candidates by similarity with
$\mathbf{b}_i$. Top-$K$ retrieval is
\begin{equation}
    \operatorname{Ret@K}
    =
    \frac{1}{N}
    \sum_{i=1}^{N}
    \mathbb{I}
    \left[
    \operatorname{rank}
    (\mathbf{g}_i \mid \mathbf{b}_i)
    \leq K
    \right],
    \label{eq:retrieval}
\end{equation}
where a lower rank is better. We report both brain-to-image
and image-to-brain retrieval~\cite{scotti2024mindeye2}.

In 2AFC, each mismatch is compared with its match:
\begin{equation}
\resizebox{0.84\columnwidth}{!}{$\displaystyle
\operatorname{2AFC}
=
\frac{1}{N(N-1)}
\sum_{i=1}^{N}
\sum_{j\neq i}
\mathbb{I}\!\left[
\operatorname{sim}(\mathbf{b}_i,\mathbf{g}_i)
>
\operatorname{sim}(\mathbf{b}_i,\mathbf{g}_j)
\right]
$}
\label{eq:twoafc}
\end{equation}
This score measures how often the true image is preferred to
a mismatched image~\cite{takagi2023high}.

\paragraph{L2: Counterfactual Description Discrimination.}
Our primary fine-grained metric is Counterfactual
Description Discrimination (CCD-hard), a ranking test inspired
by compositional vision-language evaluation such as
SugarCrepe~\cite{hsieh2023sugarcrepe}. It measures whether
the decoded representation favors the correct description
over scene-preserving but semantically incorrect
alternatives.
For image $i$, let $\mathbf{c}^{+}_i$ be the true caption
embedding and
$\mathcal{C}^{-}_i =
\{\mathbf{c}^{-}_{i,1},\ldots,
\mathbf{c}^{-}_{i,K}\}$
be a set of $K$ counterfactual alternatives. The prediction
is correct only when the true description is ranked above
every negative:
\begin{equation}
\resizebox{0.84\columnwidth}{!}{$\displaystyle
\operatorname{CCD}
=
\frac{1}{N}
\sum_{i=1}^{N}
\mathbb{I}\!\left[
\operatorname{sim}(\mathbf{b}_i,\mathbf{c}^{+}_i)
>
\max_{1\leq k\leq K}
\operatorname{sim}(\mathbf{b}_i,\mathbf{c}^{-}_{i,k})
\right]
$}
\label{eq:ccd}
\end{equation}
Unlike random-negative evaluation, this metric tests whether
the decoder can resolve a small but important conflict.

In addition to the original counterfactual set, we apply the
same CCD rule to an independent Cross-LLM Fixed-K2 set, a
Human-written Challenge, and the existing-negative set covered
by the Human Audit. These sets are fixed before model
evaluation, are shared by all variants, and are not used for
retraining. Their construction and sample counts are reported
in the Supplement.

\paragraph{L3: Representational Geometry.}
We use representational similarity analysis
(RSA)~\cite{kriegeskorte2008representational} to compare the
pairwise geometry of decoded and target representations. We
construct two $N\times N$ representational similarity
matrices, $\mathbf{S}_{brain}$ and $\mathbf{S}_{stim}$, where
each element is the cosine similarity between a pair of
samples. The primary brain--stimulus RSA score is the Pearson
correlation between their upper-triangular entries:
\begin{equation}
    \operatorname{RSA}
    =
    \rho_{\mathrm{Pearson}}
    \left(
    \operatorname{triu}(\mathbf{S}_{brain}),
    \operatorname{triu}(\mathbf{S}_{stim})
    \right).
    \label{eq:rsa}
\end{equation}
For the separate cross-subject IS-RSA analysis, we compute
Spearman correlations between corresponding decoded
representational structures. Higher values indicate closer
agreement between the compared similarity structures.

\section{Experiments}
\label{sec:experiments}

\subsection{Experimental Setup}

We evaluate ConceptAlign on the Natural Scenes Dataset
(NSD)~\cite{allen2022massive}. Full-model reconstruction
metrics are reported for Subject~01. The CCD-H, 2AFC, RSA,
and cross-subject analyses use the four commonly evaluated
subjects and their shared test images. The full model is
built on MindEye2. The negative-source study compares four
matched Subject-01 variants: counterfactual near-miss
supervision, positive-only supervision, random-caption
negatives, and CLIP-nearest negatives. All four variants use
the same data, backbone, training schedule, and evaluation
pipeline; only the negative source changes.

We report PixCorr, SSIM, AlexNet, Inception, CLIP, and SwAV
for reconstruction quality, together with CCD-H, brain-to-image
2AFC, and RSA. Complete training, software, hardware,
generation, and annotation details are provided in the
Supplement. Counterfactual captions are generated offline,
and no LLM is used at inference.

\subsection{Comparison with Existing Methods}

Table~\ref{tab:controlled_reference_final} compares
ConceptAlign with existing NSD reconstruction methods and
summarizes the controlled negative-source variants. The
comparison methods are ordered chronologically. Reconstruction
metrics are reported for Subject~01, while CCD-H, 2AFC-B2I,
and RSA summarize the four commonly evaluated subjects. The
lower block reports the Subject-01 controlled variants.

\begin{table*}[t]
\centering
\scriptsize
\setlength{\tabcolsep}{0.7pt}
\renewcommand{\arraystretch}{0.96}
\setlength{\aboverulesep}{0.25ex}
\setlength{\belowrulesep}{0.25ex}
\setlength{\cmidrulesep}{0.15ex}

\begin{tabular*}{\textwidth}{
@{\extracolsep{\fill}}
L{0.205\textwidth}
cccccccccc
@{}
}
\toprule

&
\multicolumn{4}{c}{\textbf{Low-Level}} &
\multicolumn{3}{c}{\textbf{High-Level}} &
\multicolumn{3}{c}{\textbf{Semantic Evaluation}} \\
[-1pt]

\cmidrule(lr){2-5}
\cmidrule(lr){6-8}
\cmidrule(lr){9-11}

\textbf{Method}
& \textbf{PixCorr}$\uparrow$
& \textbf{SSIM}$\uparrow$
& \textbf{Alex2}$\uparrow$
& \textbf{Alex5}$\uparrow$
& \textbf{Inc.}$\uparrow$
& \textbf{CLIP}$\uparrow$
& \textbf{SwAV}$\downarrow$
& \textbf{CCD-H}$\uparrow$
& \textbf{2AFC-B2I}$\uparrow$
& \textbf{RSA}$\uparrow$ \\
\midrule

\multicolumn{11}{@{}l}{
\textit{Comparison with existing methods}
} \\

MindEye2 (ICML 2024)
& \underline{37.40\%}
& \underline{43.90\%}
& \textbf{97.82\%}
& \underline{99.10\%}
& \underline{96.15\%}
& \underline{93.56\%}
& \underline{33.80\%}
& \underline{62.30\%}
& \textbf{98.40\%}
& \underline{28.05\%} \\

Flat-to-Round (ICCVW 2025)
& 16.50\%
& 30.50\%
& 78.20\%
& 89.00\%
& 85.10\%
& 88.30\%
& 39.80\%
& --
& --
& -- \\

MindTuner (AAAI 2025)
& 32.20\%
& 42.10\%
& 95.80\%
& 98.80\%
& 95.60\%
& \textbf{93.80\%}
& 34.00\%
& --
& --
& -- \\

ZEBRA (NeurIPS 2025)
& 13.10\%
& 37.50\%
& 74.60\%
& 81.20\%
& 72.20\%
& 71.50\%
& 50.60\%
& --
& --
& -- \\

Ours
& \textbf{39.30\%}\textsuperscript{\tiny(+1.90)}
& \textbf{44.50\%}\textsuperscript{\tiny(+0.60)}
& \underline{97.70\%}\textsuperscript{\tiny(-0.12)}
& \textbf{99.30\%}\textsuperscript{\tiny(+0.20)}
& \textbf{96.30\%}\textsuperscript{\tiny(+0.15)}
& 93.10\%\textsuperscript{\tiny(-0.70)}
& \textbf{32.00\%}\textsuperscript{\tiny(-1.80)}
& \textbf{63.65\%}\textsuperscript{\tiny(+1.35)}
& \underline{98.38\%}\textsuperscript{\tiny(-0.02)}
& \textbf{35.50\%}\textsuperscript{\tiny(+7.45)} \\

\midrule

\multicolumn{11}{@{}l}{
\textit{Ablation of negative sources}
} \\

Ours w/o counterfactual
& 27.26\%
& 22.13\%
& 49.32\%
& 43.60\%
& --
& --
& --
& 63.10\%
& 66.70\%
& 31.00\% \\

Ours + random negative
& 27.05\%
& 22.41\%
& 49.37\%
& 43.94\%
& --
& --
& --
& 62.90\%
& 67.00\%
& 29.90\% \\

Ours + CLIP-nearest
& 27.43\%
& 23.39\%
& 48.84\%
& 43.66\%
& --
& --
& --
& 62.60\%
& 66.60\%
& 28.10\% \\

\bottomrule
\end{tabular*}

\caption{
Comparison with existing methods and controlled negative
source variants on NSD in terms of three-level metrics
(\%).Bold and underlined values
denote the best and second-best results within each block,
respectively; tied best results are both bolded. Superscripts
in the Ours row report its percentage-point difference from
the strongest non-Ours result in the same column. For SwAV,
a negative difference is favorable because lower is better.
Published methods retain their original training and
evaluation settings. Complete ablation values and paired
intervals are provided in the Supplement.
}
\label{tab:controlled_reference_final}
\end{table*}
On conventional reconstruction, ConceptAlign improves
MindEye2 from 37.4\% to 39.3\% in PixCorr, from 43.9\% to
44.5\% in SSIM, and from 33.8\% to 32.0\% in SwAV. It also
improves AlexNet-5 and Inception while remaining close on
AlexNet-2 and CLIP. The results indicate that the additional
semantic objective remains compatible with strong low- and
high-level reconstruction quality.

\begin{figure}[t]
    \centering
    \includegraphics[width=0.92\linewidth]
    {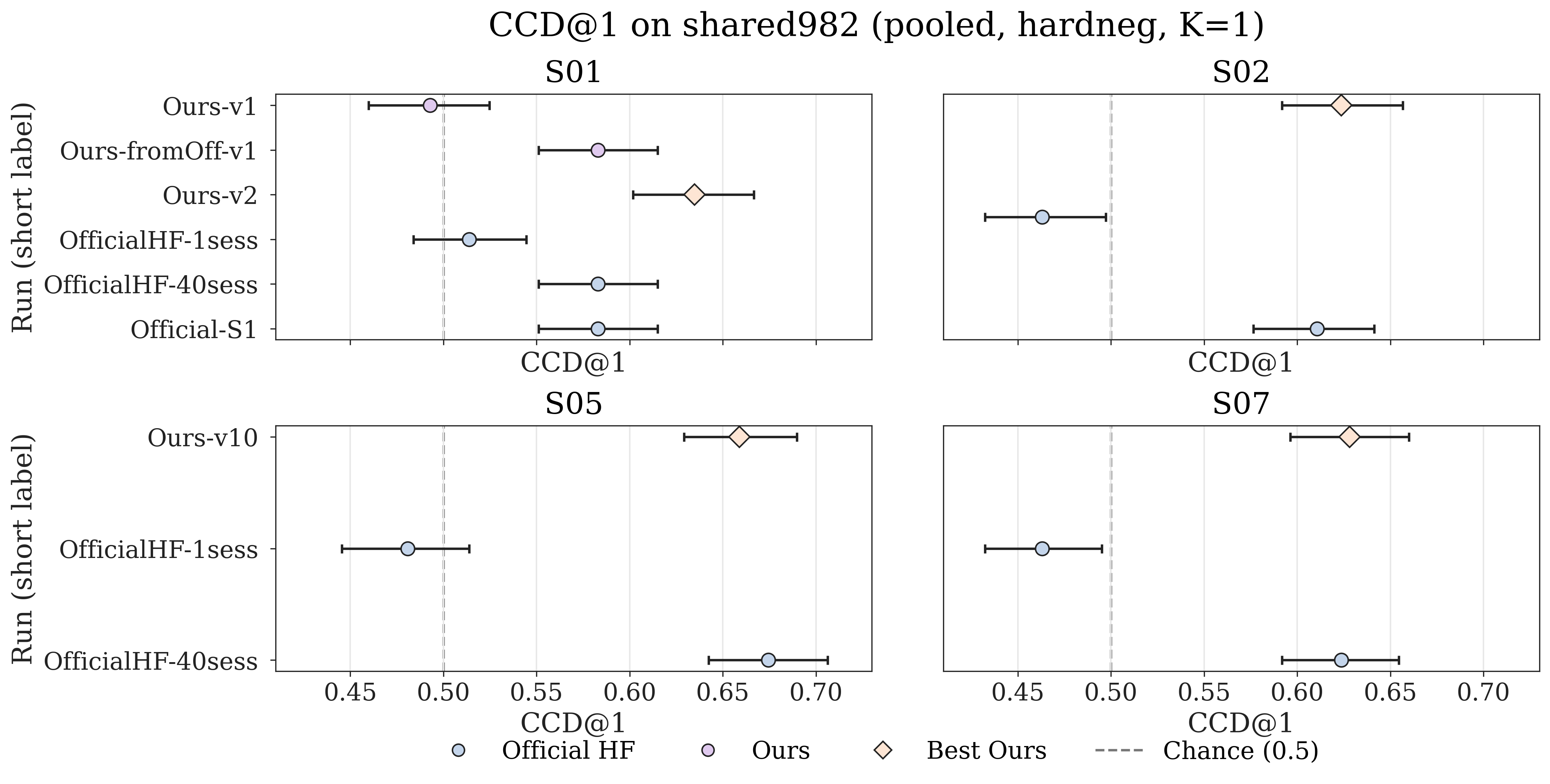}

    \vspace{2pt}

    \includegraphics[width=0.9\linewidth]
    {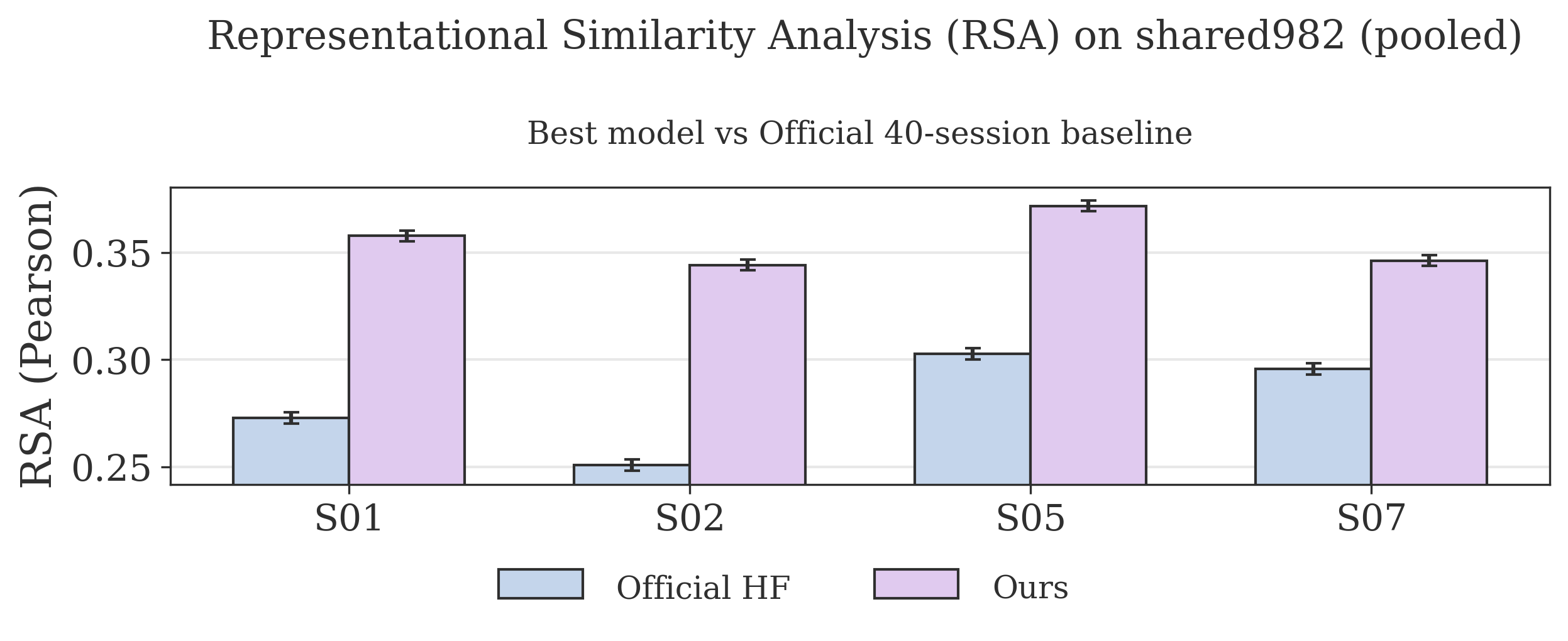}

    \caption{
    \textbf{Subject-level semantic evaluation.}
    (\textbf{Top}) CCD@1 on shared982 for the available ConceptAlign and
official MindEye2 checkpoints; error bars show 95\% bootstrap
confidence intervals. (\textbf{Bottom}) Brain--image RSA
measured by Pearson correlation. CCD comparison is diagnostic
because checkpoint availability and budgets differ across
subjects.
    }
    \label{fig:main_subject_semantics}
\end{figure}

The representation-level evaluation gives a complementary
view. CCD-H increases from 62.3\% to 63.7\%, while RSA
increases from 28.1\% to 35.5\% and 2AFC-B2I remains at
98.4\%. Figure~\ref{fig:main_subject_semantics} provides the
subject-level CCD and RSA results. Together, these results
associate counterfactual supervision with both fine-grained
semantic decisions and the broader geometry of the decoded
representation.

\subsection{Performance in Challenging Settings}
\label{sec:challenging}

\paragraph{Qualitative semantic fidelity.}
Figure~\ref{fig:semantic_fidelity} compares the two models
under anchors defined from the ground-truth images. Group A
contains cases in which both models recover the scene, but
ConceptAlign better preserves a detail, including leg pose,
window placement, cropped airplane framing, and the narrow
post below a clock. These details are small relative to the
image, yet they distinguish the observed stimulus from a
plausible completion of the same scene.

Group B contains more direct semantic differences in object
identity, group arrangement, count, and visible attributes.
For example, the anchors distinguish a bear from another
large animal, a line of people from a scattered group, and
the presence of glasses from a generic face. The examples
complement the quantitative metrics by showing which stimulus
details contribute to a semantic match, even when visual
realism makes the two reconstructions appear broadly similar.

These examples illustrate why visual quality and semantic
fidelity should be examined separately. Visual appearance
alone does not determine how faithfully a reconstruction
preserves stimulus-specific content. A strong generative
prior can produce a coherent image while introducing content
that is not supported by the
stimulus~\cite{shirakawa2025spurious}.

\paragraph{Fine-grained semantic conflicts.}
The semantic breakdown shows improvements when the negative
description remains close to the scene. CCD on object edits
increases from 58.5\% to 65.0\% for one subject and from
61.3\% to 63.9\% for another; attribute-edit performance
increases from 56.6\% to 57.7\%. Values are reported in
Supplementary
Table~\ref{tab:supp-inherited-semantic-breakdown}. These
estimates are consistent with the pattern in
Figure~\ref{fig:main_subject_semantics} and suggest that
counterfactual supervision can help when the decoder must
reject a plausible object or attribute alternative rather
than an unrelated caption.

\begin{figure}[t]
    \centering

    \begin{minipage}[c]{0.53\linewidth}
        \centering
        \includegraphics[width=\linewidth]
        {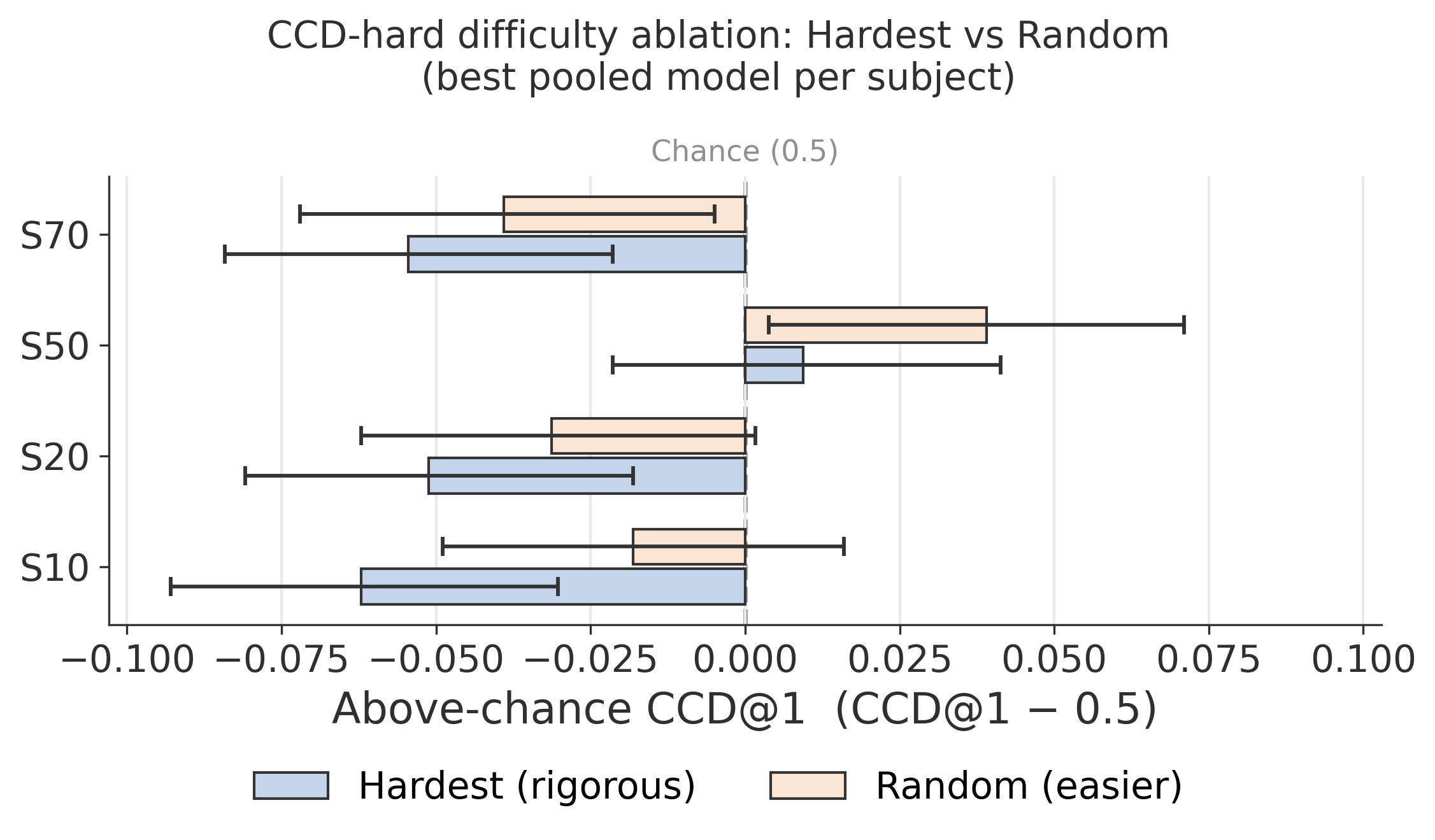}
    \end{minipage}
    \hfill
    \begin{minipage}[c]{0.43\linewidth}
        \centering
        \includegraphics[width=0.95\linewidth]
        {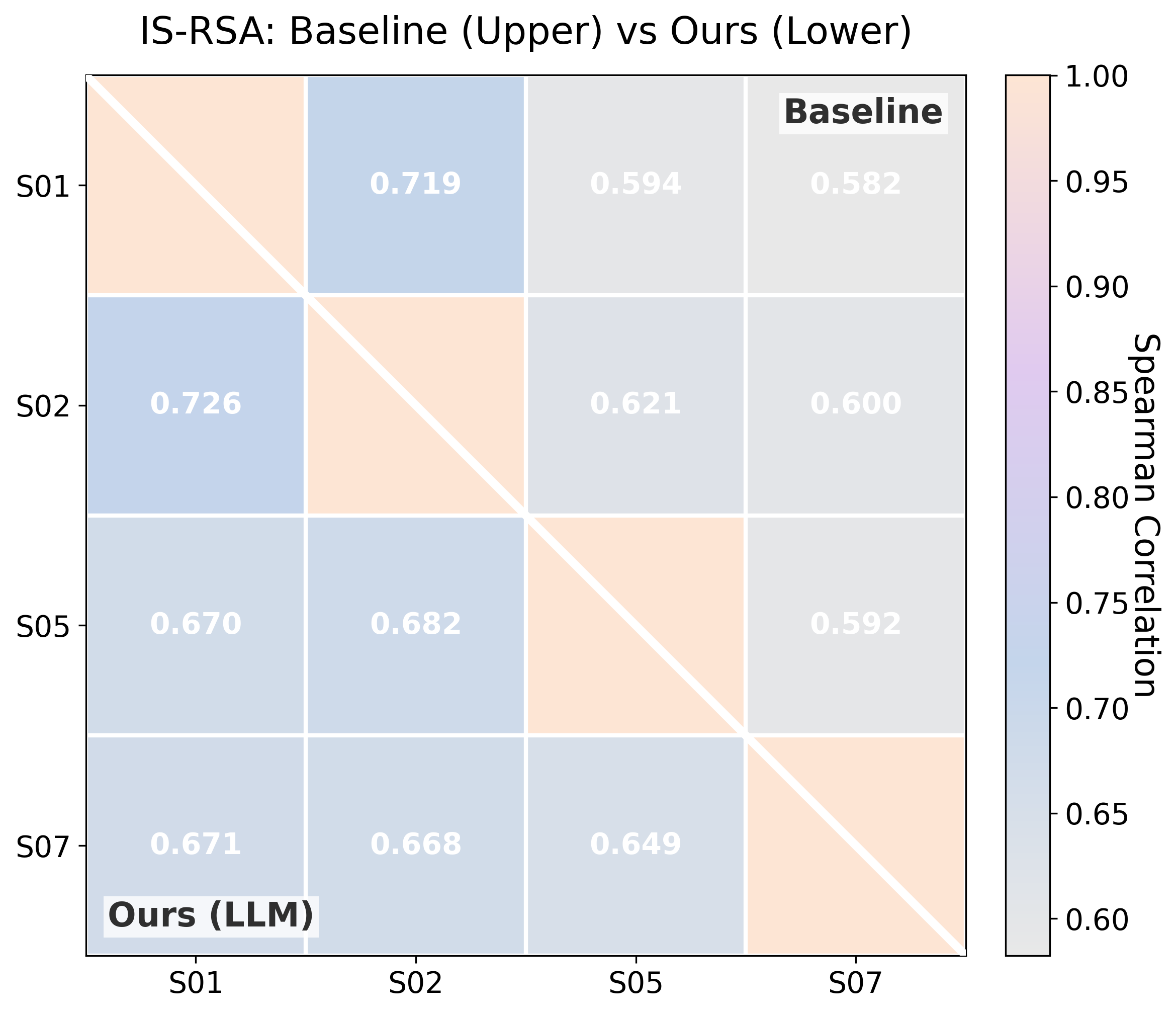}
    \end{minipage}

    \caption{
    \textbf{Semantic difficulty and cross-subject structure.}
    (\textbf{Left}) CCD@1 relative to chance under
    counterfactual and random-negative evaluation; random
    negatives define an easier task.
    (\textbf{Right}) Pairwise cross-subject IS-RSA on
    shared982. The upper triangle reports MindEye2 and the
    lower triangle reports ConceptAlign.
    }
    \label{fig:semantic_difficulty_structure}
\end{figure}

The left panel of
Figure~\ref{fig:semantic_difficulty_structure} explains why
ordinary negative evaluation is insufficient. Random
negatives produce higher scores because they are easier to
separate, whereas counterfactual negatives expose errors
hidden by coarse evaluation. The comparison therefore
characterizes semantic-test difficulty rather than a uniform
increase in accuracy.

\begin{figure}[t]
    \centering
    \includegraphics[width=0.91\linewidth]
    {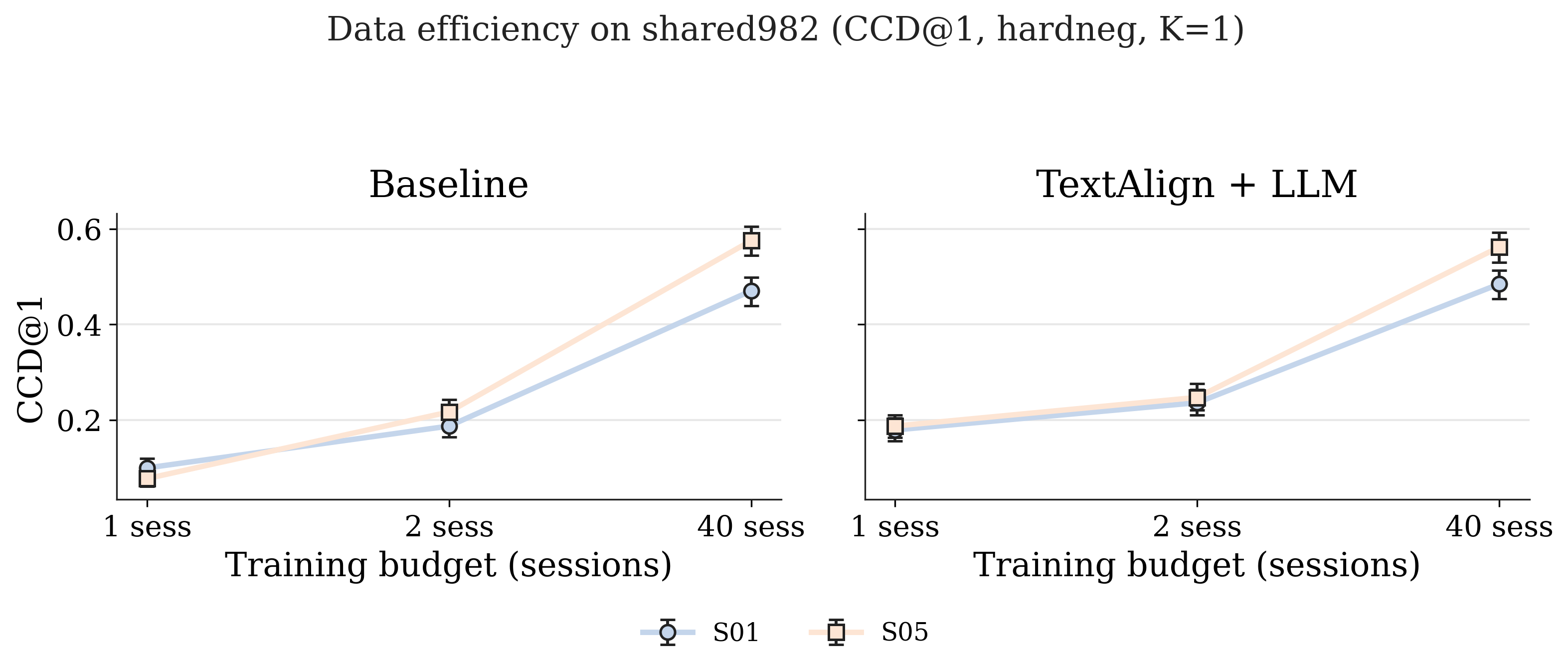}

    \vspace{2pt}

    \includegraphics[width=0.89\linewidth]
    {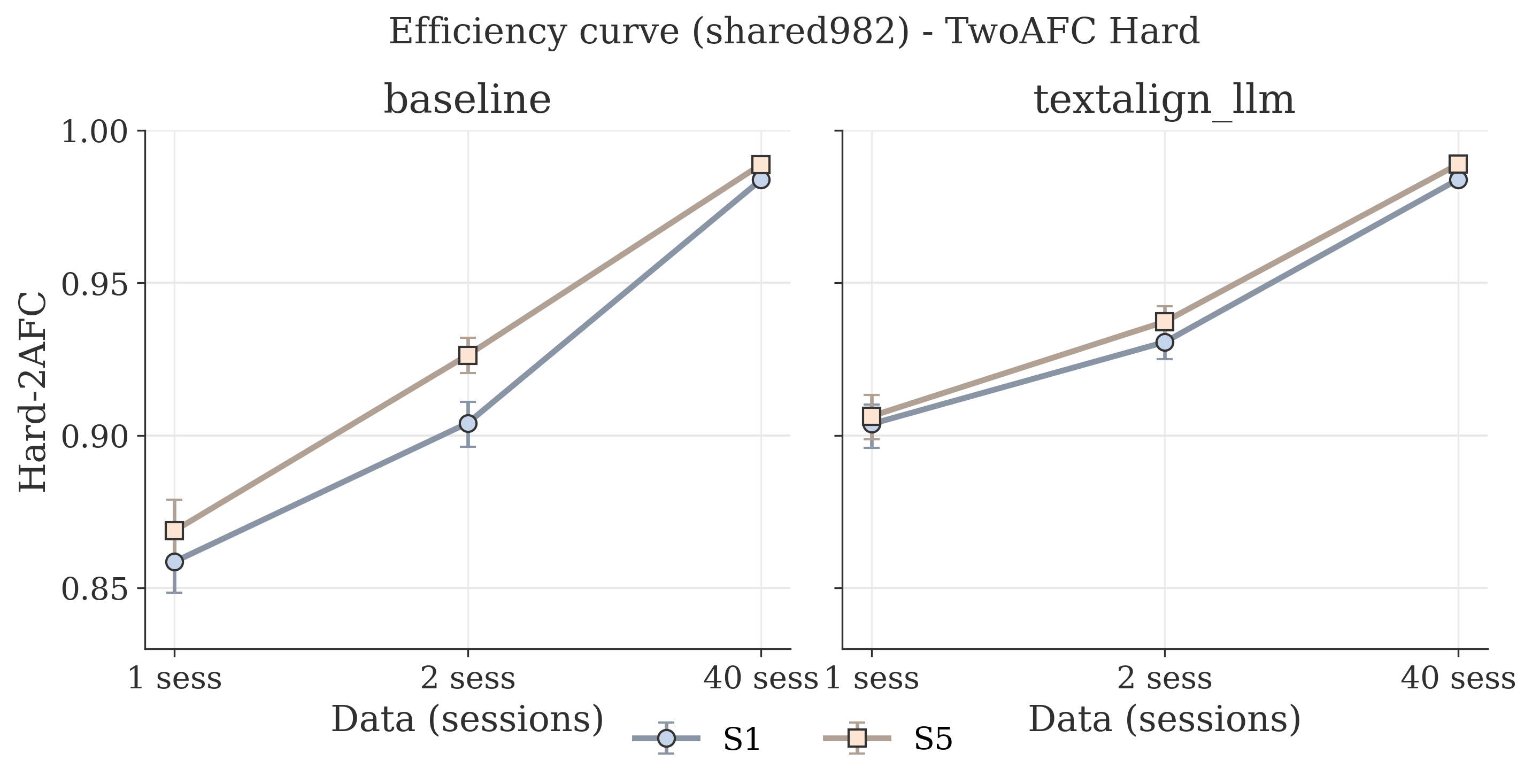}

    \caption{
    \textbf{Data efficiency with limited fMRI training data.}
    (\textbf{Top}) CCD@1 and
    (\textbf{bottom}) 2AFC-Hard for Subjects~01 and~05 using
    one, two, and forty sessions. In both evaluations, the
    ConceptAlign difference is most visible with one or two
    sessions, while the full-data results are closer.
    }
    \label{fig:main_data_efficiency}
\end{figure}

\paragraph{Limited-data decoding.} 
Figure~\ref{fig:main_data_efficiency} combines the CCD and
2AFC-Hard data-efficiency analyses for Subjects~01 and~05. In
both evaluations, the difference associated with ConceptAlign
is clearest in the one- and two-session settings, while the
full-data results become closer. With fewer fMRI samples, the
decoder receives less repeated evidence about each visual
concept. The targeted alternatives provide a semantic
constraint by specifying both the matched description and a
nearby interpretation that should be rejected. The pattern
across the two metrics suggests that this supervision remains
useful when neural training data are limited. 

\paragraph{Cross-subject structure.}
All six IS-RSA differences favor ConceptAlign, ranging from
+.057 to +.089. As shown in the right panel of
Figure~\ref{fig:semantic_difficulty_structure}, the decoded
representations preserve a more similar relational structure
across subjects. The subject-wise brain--image RSA results in
Figure~\ref{fig:main_subject_semantics} provide a
complementary comparison with the target-image geometry.

\subsection{Ablation and Analysis}

\paragraph{Contribution of counterfactual supervision.}
The lower block of
Table~\ref{tab:controlled_reference_final} compares
counterfactual near-miss supervision with positive-only,
random-caption, and CLIP-nearest alternatives. ConceptAlign
retains the strongest 2AFC-B2I and RSA results while also
improving PixCorr and both AlexNet measures over the
controlled variants.

Table~\ref{tab:main_ablation_details} summarizes the
negative-source characteristics and ranking-margin
sensitivity. Random captions are substantially farther from
the positive caption, whereas CLIP-nearest and counterfactual
negatives have similar embedding proximity. Among the tested
margins, $m=.1$ gives the highest CCD@1 and ties the strongest
retrieval result.

\begin{table}[t]
\centering
\scriptsize
\setlength{\tabcolsep}{2.1pt}
\renewcommand{\arraystretch}{1.03}

\begin{tabular*}{\columnwidth}{
@{\extracolsep{\fill}}
llcccc
@{}
}
\toprule
\multicolumn{3}{c}{\textbf{Negative source}} &
\multicolumn{3}{c}{\textbf{Margin sensitivity}} \\
\cmidrule(lr){1-3}
\cmidrule(lr){4-6}

\textbf{Variant} &
\textbf{Source} &
\textbf{Cos. (\%)} &
\textbf{$m$} &
\textbf{CCD@1 (\%)} &
\textbf{Ret.@1 (\%)} \\
\midrule

Positive-only
& None
& --
& .05
& 59.20
& 17.00 \\

Random
& Random COCO
& 34.83
& \textbf{.10}
& \textbf{63.60}
& \textbf{18.00} \\

CLIP-NN
& Nearest CLIP
& 81.62
& .20
& 62.10
& \textbf{18.00} \\

ConceptAlign
& CF near-miss
& $82.00\pm5.00$
& .50
& 61.00
& 16.00 \\

\bottomrule
\end{tabular*}

\caption{
Negative-source characteristics and margin sensitivity.
Cos. denotes caption-to-positive cosine similarity. All
percentage-valued columns are reported in percent. Ret.@1
denotes forward retrieval. The two table blocks are
independent; rows are aligned only for compact presentation.
}
\label{tab:main_ablation_details}
\end{table}

\begin{table}[t]
\centering
\scriptsize
\setlength{\tabcolsep}{0.8pt}
\renewcommand{\arraystretch}{1.06}

\begin{tabular}{
@{}
L{0.25\columnwidth}
C{0.06\columnwidth}
C{0.09\columnwidth}
C{0.09\columnwidth}
C{0.13\columnwidth}
C{0.21\columnwidth}
@{}
}
\toprule

\multicolumn{6}{c}{
\textbf{Human audit} ($N=534$)
} \\
\midrule

\multicolumn{2}{l}{Scene preserved (\%)}
& 97.90
& \multicolumn{2}{l}{Near-miss score (1--5)}
& 3.55 \\

\multicolumn{2}{l}{Explicit negation (\%)}
& 0.20
& \multicolumn{2}{l}{Harmful/bias (\%)}
& 0.00 \\

\midrule

\multicolumn{6}{c}{
\textbf{Independent evaluation} (Acc@1, \%)
} \\
\midrule

\textbf{Set}
& \textbf{N}
& \textbf{Ours}
& \textbf{Ref.}
& \textbf{$\Delta$ (pp)}
& \textbf{95\% CI (pp)} \\
\midrule

Cross-LLM
& 979
& 35.24
& 36.06
& -0.82
& $[-3.17,1.53]$ \\

Human-written
& 100
& 58.00
& 61.00
& -3.00
& $[-11.00,4.00]$ \\

Human-audited
& 178
& \textbf{78.09}
& 71.91
& \textbf{+6.18}
& $\mathbf{[1.12,11.24]}$ \\

\bottomrule
\end{tabular}

\caption{
Human audit and independent counterfactual evaluation.
Human-audit rates and Acc@1 are reported in percent;
differences and confidence intervals in percentage points
(pp). Ref. denotes the strongest matched control for
Cross-LLM and Human-written evaluation and CLIP-nearest for
the Human-audited set. Scene preservation denotes the
percentage of audited negatives judged to preserve the scene.
}
\label{tab:human_audit_final}
\end{table}

\paragraph{Counterfactual quality and human tests.}
Table~\ref{tab:human_audit_final} summarizes the human audit
and evaluation on independent counterfactual sources. The
audit shows high scene preservation, moderate near-miss
plausibility, little explicit negation, and no harmful or
biased content.

The paired intervals include zero on the Cross-LLM and
Human-written sets. On the Human-audited set, ConceptAlign
exceeds CLIP-nearest by .062, with a 95\% interval of
$[.011,.112]$. Per-model results, semantic-error ratings, and
annotator agreement are reported in the Supplement.
\section{Conclusion}

We introduced ConceptAlign, a semantic alignment module that
trains a visual brain decoder to distinguish the observed
scene from plausible counterfactual alternatives. On
Subject~01, ConceptAlign improves its MindEye2 backbone on
several reconstruction measures. Across the four subjects,
the semantic analyses show higher CCD-H and representational
alignment. The matched Subject-01 ablation also shows
benefits in 2AFC-B2I, RSA, PixCorr, and AlexNet features.

The qualitative examples further show that semantic fidelity
cannot be judged from visual realism alone. Human auditing,
Cross-LLM evaluation, and human-written alternatives support
the quality and transfer of the counterfactual supervision.
Overall, visual brain decoding benefits not only from
matching what was observed, but also from learning which
plausible alternatives conflict with the stimulus.
\section*{Ethical Statement}

This study uses the de-identified Natural Scenes Dataset
under its established consent and data-use procedures. The
proposed method is intended for research on visual
representation and reconstruction rather than for clinical,
diagnostic, or surveillance applications.

Counterfactual captions are generated offline using
constrained prompts and fixed lexical and embedding-based
filters. The LLM-generated training captions, the independent
Cross-LLM evaluation set, and the Human-written Challenge are
kept separate according to their designated roles. The
Human-written Challenge and Human Audit are used only for
evaluation and do not provide training supervision. Human
annotators assess scene preservation, semantic plausibility,
explicit negation, and potentially harmful or biased content.
No LLM or human annotation is required during inference.

\bibliography{ijcai26}

\appendix

\renewcommand{\thesection}{\Alph{section}}
\setcounter{section}{0}
\setcounter{figure}{0}
\setcounter{table}{0}
\setcounter{equation}{0}

\renewcommand{\thefigure}{\thesection\arabic{figure}}
\renewcommand{\thetable}{\thesection\arabic{table}}
\renewcommand{\theequation}{\thesection\arabic{equation}}

\raggedbottom
\setcounter{topnumber}{5}
\setcounter{bottomnumber}{5}
\setcounter{totalnumber}{10}
\setcounter{dbltopnumber}{3}
\renewcommand{\topfraction}{0.98}
\renewcommand{\bottomfraction}{0.95}
\renewcommand{\textfraction}{0.01}
\renewcommand{\floatpagefraction}{0.92}
\renewcommand{\dbltopfraction}{0.98}
\renewcommand{\dblfloatpagefraction}{0.92}
\setlength{\textfloatsep}{8pt plus 2pt minus 2pt}
\setlength{\floatsep}{7pt plus 2pt minus 2pt}
\setlength{\intextsep}{7pt plus 2pt minus 2pt}
\setlength{\dbltextfloatsep}{8pt plus 2pt minus 2pt}
\setlength{\dblfloatsep}{7pt plus 2pt minus 2pt}

\twocolumn[
\begin{center}
{\LARGE\bfseries From ``What-If'' to ``What-Is'':\par}
{\LARGE\bfseries Counterfactual Thinking-Inspired Semantic Alignment\par}
{\LARGE\bfseries for Visual Brain Decoding\par}
\vspace{3pt}
{\LARGE\bfseries Supplementary Material\par}
\end{center}
\vspace{0.6em}
]

\section*{Overview}
Appendix~A describes counterfactual-caption construction and
the human-evaluation protocols. Appendix~B reports the
controlled negative-source study and additional results for
retrieval, 2AFC, CCD-H, RSA, limited-data decoding, and
qualitative examples. Appendix~C reports the margin analysis,
and Appendix~D gives the experimental settings.

\section{Counterfactual Construction and Human Protocols}
\label{supp:impl}

\subsection{Training-Time Counterfactual Construction}

Counterfactual captions are generated offline from canonical
COCO captions. For each positive caption, the language model
is asked to generate 12 alternatives: four object edits, four
attribute edits, and four relation edits. Each alternative is
required to preserve the overall scene while changing one
visual fact. The returned captions are parsed, screened,
encoded in the frozen teacher text space, and ranked by
cosine similarity to the positive caption. No score from an
evaluated decoder is used during construction.

The controlled fine-tuning set contains 8,946 fMRI training
samples. Among them, 909 captioned stimulus entries provide
counterfactual text supervision. All samples continue to
contribute to the inherited retrieval, diffusion-prior, and
low-level reconstruction objectives.

\begin{algorithm}[!tb]
\caption{Offline Construction of Counterfactual Negatives}
\label{alg:hardneg}
\begin{algorithmic}[1]
\STATE \textbf{Input:} Positive caption $C_{gt}$, aligned
positive text feature $\mathbf{t}_{pos}$, language model, and
frozen text encoder $\mathcal{E}_{text}$.
\STATE Request 12 candidates: four object edits, four
attribute edits, and four relation edits.
\FOR{$a=1,\ldots,6$}
    \STATE Parse the JSON response into candidate records.
    \STATE Retain records with a valid edit type and a
    non-empty counterfactual caption.
    \STATE Apply the fixed format and lexical filters used in
    the main paper.
    \STATE Retain captions whose word-count ratio to
    $C_{gt}$ lies in $[0.70,1.30]$.
    \STATE Retain captions whose word-set Jaccard overlap
    with $C_{gt}$ is at least $0.20$.
    \IF{at least six candidates remain}
        \STATE \textbf{break}
    \ENDIF
\ENDFOR
\IF{fewer than six candidates remain after all attempts}
    \STATE Mark the sample as invalid for generated-negative
    supervision.
\ELSE
    \FOR{each retained caption $c_j^{-}$}
        \STATE
        $\mathbf{u}_{j}^{-}
        \leftarrow
        \operatorname{Norm}
        \bigl(\mathcal{E}_{text}(c_j^{-})\bigr)$.
        \STATE
        $s_j
        \leftarrow
        \langle
        \mathbf{t}_{pos},
        \mathbf{u}_{j}^{-}
        \rangle$.
    \ENDFOR
    \STATE Use the fixed similarity range $[0.15,0.85]$ to
    rank the screened candidates.
    \STATE Store the highest-ranked selected embedding.
\ENDIF
\end{algorithmic}
\end{algorithm}

\subsection{Training Prompt Template}

The following prompt is used for training-caption generation.
The Cross-LLM evaluation set uses a separate prompt and
generator.

\paragraph{System prompt.}
\begin{quote}
\footnotesize\ttfamily\raggedright
You are a careful assistant. Output JSON only.
\end{quote}

\paragraph{User prompt template.}
\begin{quote}
\scriptsize\ttfamily\raggedright

You are helping me generate stimulus-preserving
counterfactual hard-negative captions for contrastive
learning in brain decoding.

\medskip

Given ONE ground-truth caption describing an image, generate
a set of counterfactual captions that are:

\medskip

(1) stimulus-preserving: they must still describe the SAME
overall scene as the ground-truth (same setting, viewpoint,
action, and global context);

\medskip

(2) critically wrong: each caption must contain EXACTLY ONE
targeted semantic deviation that would make it incorrect for
the image;

\medskip

(3) near-miss: the deviation should be subtle and plausible
(hard negative), not obviously wrong or absurd;

\medskip

(4) implicit: DO NOT use explicit negation or correction
language (forbidden words include: ``not'', ``no'', ``never'',
``instead'', ``wrong'', ``but'', ``however'', ``rather than'',
and ``without''). Phrase it as an alternative hypothesis,
like a natural caption.

\medskip

You must generate 12 candidates total with the following
coverage:

\medskip

- 4 Object counterfactuals: change the identity/category of
ONE main object to a visually plausible alternative while
keeping everything else consistent.

\medskip

- 4 Attribute counterfactuals: change ONE attribute of ONE
object (color, material, number, size, or age) while keeping
object identity and scene consistent.

\medskip

- 4 Relation counterfactuals: change ONE relationship,
spatial relation, or action relation while keeping the
objects and setting consistent.

\medskip

Hardness constraints:

\medskip

- Make the counterfactual captions highly similar in wording
to the ground-truth caption. Paraphrase minimally and keep
most tokens unchanged.

\medskip

- Keep the sentence length and style similar to the
ground-truth caption.

\medskip

- Keep the same number of main entities as much as possible;
do not introduce new objects unless replacing exactly one
object.

\medskip

- Preserve the global scene semantics, including place,
activity, viewpoint, and background.

\medskip

- Each candidate must differ from the ground-truth by ONE and
ONLY ONE key semantic edit. Avoid multiple edits.

\medskip

- Avoid trivial edits that do not change meaning.

\medskip

Return the result as STRICT JSON with no markdown, as a list
of 12 items. Each item must contain:

\medskip

\{
``type'': ``object'' $|$ ``attribute'' $|$ ``relation'',

``edit'': ``a short description of the changed detail'',

``caption\_cf'': ``the counterfactual caption'',

``changed\_span'': ``the changed phrase in caption\_cf''
\}

\medskip

Ground-truth caption:

``\{pos\_caption\}''
\end{quote}

\subsection{Caption Similarity}

The 909 captioned entries used for text supervision have a
mean positive-to-counterfactual cosine similarity of
$82.00\%\pm5.00\%$. Explicit negation occurs in 0.00\% of
the retained entries. Figure~\ref{fig:audit} shows the
similarity distribution.

\par\medskip
\noindent
\begin{minipage}{\columnwidth}
    \centering
    \includegraphics[width=\linewidth]
    {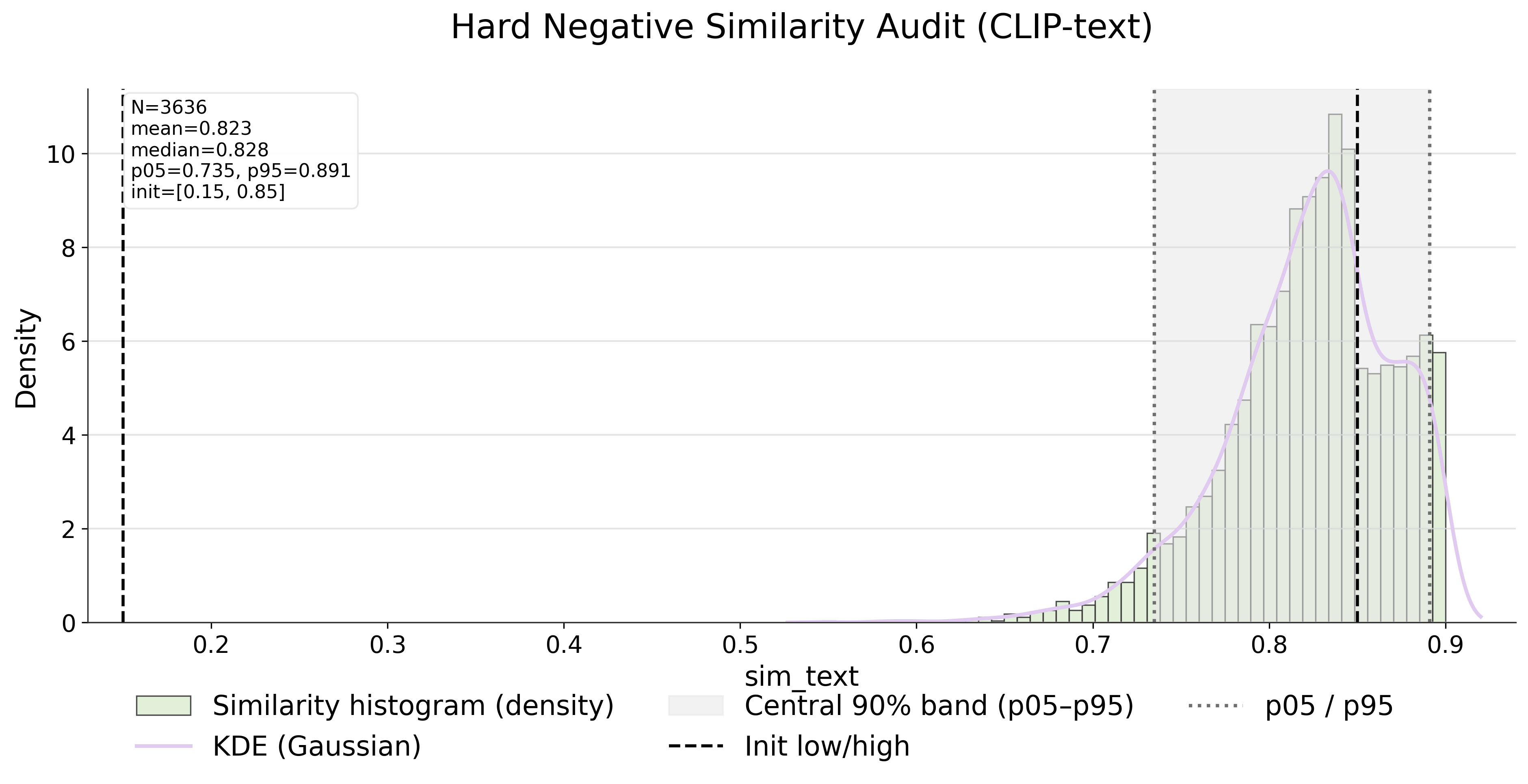}
    \captionof{figure}{Cosine similarity between positive
    captions and retained counterfactual alternatives in the
    caption-supervised subset.}
    \label{fig:audit}
\end{minipage}
\par\medskip

\subsection{Cross-LLM Fixed-K2 Protocol}
\label{supp:crossllm-protocol}

The final Cross-LLM set contains 979 images with two
counterfactual negatives per image. For each source image,
GPT-5.4-mini generates object, attribute, and relation edits,
and frozen OpenCLIP filtering is applied without access to
model scores. The Cross-LLM, Human-written, and Human-audited
sets are fixed before evaluation and used only for testing.

\subsection{Human-Written Counterfactual Challenge}

Two writers provide one counterfactual caption for each of
100 images, producing 200 entries. A reviewer checks the
captions and replaces those that do not follow the task
instructions. The final set contains 190 writer originals and
10 reviewer corrections. All captions are represented in
English for text-encoder evaluation.

\subsection{Human Audit Protocol}

Three annotators rate 534 existing LLM-generated negatives
from 178 images, giving 1,602 ratings. The dimensions are
scene preservation, semantic plausibility, near-miss
plausibility, edit type, explicit negation, and harmful or
biased content. Near-miss plausibility is rated on a
five-point scale, with higher scores indicating a more
plausible near miss. The annotators assess existing captions
and do not provide training supervision.

\begin{table}[!tb]
\centering
\footnotesize
\setlength{\tabcolsep}{3pt}
\renewcommand{\arraystretch}{1.05}
\begin{tabular}{@{}L{0.39\columnwidth}L{0.52\columnwidth}@{}}
\toprule
\textbf{Item} & \textbf{Value} \\
\midrule
Negatives / images & 534 / 178 \\
Annotators & 3 \\
Total ratings & 1,602 \\
Task & Rate each caption by dimension \\
Use in training & No \\
\bottomrule
\end{tabular}
\caption{Human Audit protocol. Results are reported for each
rating dimension.}
\label{tab:supp-human-audit-protocol}
\end{table}

\section{Additional Experimental Results}
\label{supp:results}

\subsection{Controlled Negative-Source Study}

The Subject-01 study compares ConceptAlign with three
controls under the same data, backbone, training budget, and
evaluation protocol. C1 removes counterfactual supervision,
C2 uses random COCO captions, and C3 uses CLIP-nearest
captions. Table~\ref{tab:supp-controlled-reconstruction}
reports the reconstruction results, and
Table~\ref{tab:negative_sources} summarizes the negative
sources.

\begin{table}[!tb]
\centering
\footnotesize
\setlength{\tabcolsep}{2.8pt}
\begin{tabular*}{\columnwidth}{@{\extracolsep{\fill}}lrrrr@{}}
\toprule
\textbf{Metric (\%)} & \textbf{Ours} & \textbf{C1} &
\textbf{C2} & \textbf{C3} \\
\midrule
PixCorr & \textbf{39.30} & 27.26 & 27.05 & 27.43 \\
SSIM & \textbf{44.50} & 22.13 & 22.41 & 23.39 \\
AlexNet L2 & \textbf{97.70} & 49.32 & 49.37 & 48.84 \\
AlexNet L5 & \textbf{99.30} & 43.60 & 43.94 & 43.66 \\
\bottomrule
\end{tabular*}
\caption{Reconstruction results for the controlled
negative-source study on Subject~01 ($N=982$). All variants
are evaluated with the same metric implementation.}
\label{tab:supp-controlled-reconstruction}
\end{table}

\begin{table}[!tb]
\centering
\footnotesize
\setlength{\tabcolsep}{3pt}
\renewcommand{\arraystretch}{1.03}
\begin{tabular}{@{}
L{0.34\columnwidth}
L{0.39\columnwidth}
C{0.19\columnwidth}
@{}}
\toprule
\textbf{Variant} &
\textbf{Negative source} &
\textbf{Cosine (\%)} \\
\midrule
C1: Ours w/o counterfactual & None & -- \\
C2: Ours + random negative & Random COCO caption & 34.83 \\
C3: Ours + CLIP-nearest & Nearest CLIP caption & 81.62 \\
Ours (ConceptAlign) & Counterfactual near-miss & $82.00\pm5.00$ \\
\bottomrule
\end{tabular}
\caption{Negative sources used in the controlled study.
Cosine similarity is multiplied by 100 for presentation.}
\label{tab:negative_sources}
\end{table}

\paragraph{Paired confidence intervals.}
Table~\ref{tab:supp-controlled-delta-semantic} reports paired
intervals for CCD-H and brain-to-image 2AFC. Each difference
is Ours minus the stated variant. Intervals use 10,000 paired
image-level bootstrap resamples with seed 2027.

\begin{table}[!tb]
\centering
\scriptsize
\setlength{\tabcolsep}{2.2pt}
\resizebox{\columnwidth}{!}{%
\begin{tabular}{@{}llrrr@{}}
\toprule
\textbf{Metric} & \textbf{Comparison} &
\textbf{$\Delta$ (pp)} & \textbf{CI low (pp)} &
\textbf{CI high (pp)} \\
\midrule
CCD-H & Ours--C1 & -0.34 & -2.69 & 2.13 \\
CCD-H & Ours--C2 & -0.11 & -2.58 & 2.36 \\
CCD-H & Ours--C3 & 0.11 & -2.36 & 2.47 \\
2AFC B2I & Ours--C1 & 0.61 & -1.12 & 2.34 \\
2AFC B2I & Ours--C2 & 0.31 & -1.43 & 2.14 \\
2AFC B2I & Ours--C3 & 0.71 & -1.02 & 2.55 \\
\bottomrule
\end{tabular}%
}
\caption{Paired intervals for CCD-H and brain-to-image 2AFC
in the controlled negative-source study.}
\label{tab:supp-controlled-delta-semantic}
\end{table}

\subsection{L1: Foundational Discriminability}

Tables~\ref{tab:cross_retrieval} and~\ref{tab:2afc_cross}
report retrieval and 2AFC on the 982 shared test images.
Image-to-brain 2AFC is higher for ConceptAlign on all four
subjects, while brain-to-image 2AFC is near ceiling for both
methods. Retrieval varies by subject and direction.

\begin{table}[!tb]
\centering
\scriptsize
\setlength{\tabcolsep}{1.8pt}
\begin{tabular*}{\columnwidth}{@{\extracolsep{\fill}}clrrrr@{}}
\toprule
\textbf{S} & \textbf{Model} & \textbf{F@1 (\%)} &
\textbf{F@5 (\%)} & \textbf{B@1 (\%)} &
\textbf{B@5 (\%)} \\
\midrule
1 & MindEye2 & 15.00 & 41.00 & 5.00 & 13.00 \\
1 & Ours & 18.00 & 45.00 & 8.00 & 26.00 \\
\midrule
2 & MindEye2 & 16.00 & 41.00 & 4.00 & 14.00 \\
2 & Ours & 14.00 & 41.00 & 7.00 & 23.00 \\
\midrule
5 & MindEye2 & 38.00 & 74.00 & 22.00 & 52.00 \\
5 & Ours & 21.00 & 54.00 & 16.00 & 42.00 \\
\midrule
7 & MindEye2 & 24.00 & 54.00 & 12.00 & 31.00 \\
7 & Ours & 12.00 & 41.00 & 6.00 & 24.00 \\
\bottomrule
\end{tabular*}
\caption{Multi-subject pooled retrieval on the 982 shared
test images. F and B denote forward and backward retrieval.
Values are reported in percent.}
\label{tab:cross_retrieval}
\end{table}

\begin{table}[!tb]
\centering
\footnotesize
\setlength{\tabcolsep}{3pt}
\begin{tabular*}{\columnwidth}{@{\extracolsep{\fill}}clcc@{}}
\toprule
\textbf{Subj} & \textbf{Model} &
\textbf{B$\rightarrow$I (\%)} &
\textbf{I$\rightarrow$B (\%)} \\
\midrule
1 & MindEye2 & 97.80 & 84.70 \\
1 & Ours & 98.40 & 96.40 \\
\midrule
2 & MindEye2 & 97.70 & 85.00 \\
2 & Ours & 98.10 & 96.10 \\
\midrule
5 & MindEye2 & 99.50 & 98.10 \\
5 & Ours & 98.90 & 98.30 \\
\midrule
7 & MindEye2 & 98.60 & 94.90 \\
7 & Ours & 98.10 & 96.30 \\
\bottomrule
\end{tabular*}
\caption{Multi-subject 2AFC results on the 982 shared test
images. Values are reported in percent.}
\label{tab:2afc_cross}
\end{table}

\subsection{L2: Counterfactual Description Discrimination}

Table~\ref{tab:supp-semantic-breakdown} reports the
object- and attribute-edit results. The largest differences
occur for object edits in Subjects~01 and~02.

\begin{table}[!tb]
\centering
\footnotesize
\setlength{\tabcolsep}{2.4pt}
\begin{tabular*}{\columnwidth}{@{\extracolsep{\fill}}lcrrc@{}}
\toprule
\textbf{Type (N)} & \textbf{S} & \textbf{MindEye2 (\%)} &
\textbf{Ours (\%)} & \textbf{Higher} \\
\midrule
Object (682) & 1 & 58.50 & \textbf{65.00} & Ours \\
Object (682) & 2 & 61.30 & \textbf{63.90} & Ours \\
Attribute (182) & 1 & 56.60 & \textbf{57.70} & Ours \\
\bottomrule
\end{tabular*}
\caption{CCD results by semantic edit type. Values are
reported in percent.}
\label{tab:supp-semantic-breakdown}
\label{tab:supp-inherited-semantic-breakdown}
\end{table}

\subsubsection{Independent Counterfactual Evaluation}

The paired intervals include zero for the Cross-LLM and
Human-written sets. On the Human-audited set, ConceptAlign
exceeds the CLIP-nearest control by 6.18 percentage points,
with a 95\% interval of $[1.12,11.24]$.
Tables~\ref{tab:supp-external-absolute} and
\ref{tab:supp-external-deltas} report the per-variant results
and paired comparisons.

\begin{table}[!tb]
\centering
\scriptsize
\setlength{\tabcolsep}{2.2pt}
\begin{tabular*}{\columnwidth}{@{\extracolsep{\fill}}lcrrrr@{}}
\toprule
\textbf{Set} & \textbf{N} & \textbf{Ours (\%)} &
\textbf{C1 (\%)} & \textbf{C2 (\%)} & \textbf{C3 (\%)} \\
\midrule
Cross-LLM K2 & 979 & 35.24 & 35.04 & \textbf{36.06} & 35.75 \\
Human-written & 100 & 58.00 & 59.00 & 59.00 & \textbf{61.00} \\
Human-audited & 178 & \textbf{78.09} & 74.72 &
\textbf{78.09} & 71.91 \\
\bottomrule
\end{tabular*}
\caption{Image-level Acc@1 (\%) on the three evaluation-only
counterfactual sets.}
\label{tab:supp-external-absolute}
\end{table}

\begin{table}[!tb]
\centering
\scriptsize
\setlength{\tabcolsep}{1.8pt}
\resizebox{\columnwidth}{!}{%
\begin{tabular}{@{}llrrr@{}}
\toprule
\textbf{Set} & \textbf{Comparison} &
\textbf{$\Delta$ (pp)} & \textbf{CI low (pp)} &
\textbf{CI high (pp)} \\
\midrule
Cross-LLM & Ours--C1 & +0.20 & -2.04 & +2.35 \\
Cross-LLM & Ours--C2 & -0.82 & -3.17 & +1.53 \\
Cross-LLM & Ours--C3 & -0.51 & -3.47 & +2.35 \\
Human-written & Ours--C1 & -1.00 & -7.00 & +5.00 \\
Human-written & Ours--C2 & -1.00 & -7.00 & +5.00 \\
Human-written & Ours--C3 & -3.00 & -11.00 & +4.00 \\
Human-audited & Ours--C1 & +3.37 & -1.69 & +8.43 \\
Human-audited & Ours--C2 & +0.00 & -5.06 & +5.06 \\
Human-audited & Ours--C3 & +6.18 & +1.12 & +11.24 \\
\bottomrule
\end{tabular}%
}
\caption{Paired Acc@1 differences. $\Delta$ is Ours minus the
stated variant. Intervals use 10,000 image-level bootstrap
resamples with seed 2027.}
\label{tab:supp-external-deltas}
\end{table}

\subsubsection{Human Audit Results}

Table~\ref{tab:supp-human-audit-results} reports the audit
outcomes. Most captions preserve the scene, explicit negation
is rare, and no harmful or biased content is flagged.
Semantic plausibility records whether the replacement is
incoherent or unreasonable.

\begin{table}[!tb]
\centering
\footnotesize
\setlength{\tabcolsep}{2.7pt}
\begin{tabular*}{\columnwidth}{
@{\extracolsep{\fill}}
lcc
@{}}
\toprule
\textbf{Dimension} & \textbf{Result} & \textbf{N} \\
\midrule
Scene preserved & 97.90\% & 534 \\
Implausible replacement (majority) & 6.90\% & 533 \\
Near-miss mean / median & 3.55 / 3.67 & 534 \\
Invalid edit type & 0.00\% & 527 \\
Other edit type & 0.00\% & 527 \\
Explicit negation & 0.20\% & 534 \\
Harmful/bias flag & 0.00\% & 534 \\
\bottomrule
\end{tabular*}
\caption{Human Audit outcomes by dimension. Seven edit-type
items were undetermined.}
\label{tab:supp-human-audit-results}
\end{table}

Table~\ref{tab:supp-human-audit-agreement} reports raw
agreement for the categorical ratings.

\begin{table}[!tb]
\centering
\scriptsize
\setlength{\tabcolsep}{2pt}
\begin{tabular*}{\columnwidth}{@{\extracolsep{\fill}}lcc@{}}
\toprule
\textbf{Dimension} & \textbf{Pairwise (\%)} &
\textbf{Unanimous (\%)} \\
\midrule
Scene preservation & 83.35 & 75.00 \\
Semantic plausibility & 79.49 & 69.25 \\
Edit type & 68.59 & 53.48 \\
No explicit negation & 98.49 & 97.74 \\
Harmful/bias flag & 99.24 & 98.85 \\
\bottomrule
\end{tabular*}
\caption{Raw agreement for the categorical Human Audit
ratings. Pairwise agreement is the proportion of equal
observed rating pairs; unanimous agreement requires all three
ratings to match.}
\label{tab:supp-human-audit-agreement}
\end{table}

\subsection{L3: Representational Geometry}

\begin{figure*}[!t]
    \centering
    \includegraphics[width=\textwidth]
    {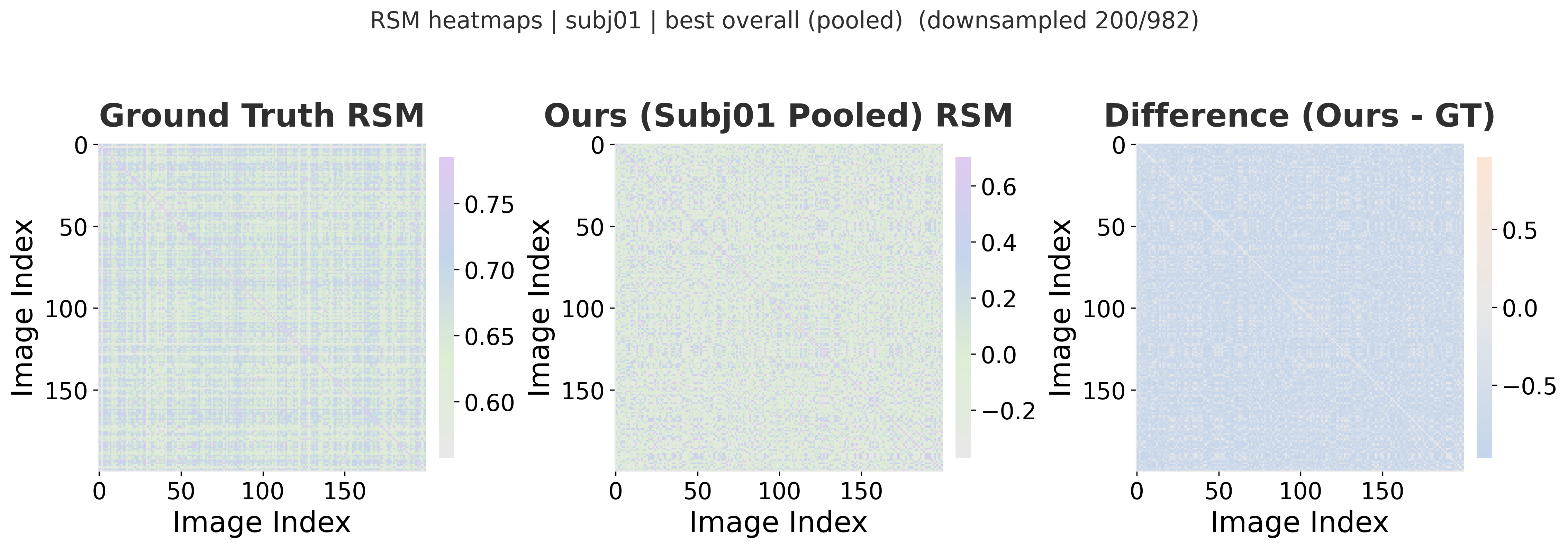}
    \caption{Representational similarity matrices for
    Subject~01 using 200 images from the shared test set.
    Left: target-image RSM. Middle: ConceptAlign-decoded RSM.
    Right: element-wise difference. The panels use separate
    color ranges and illustrate the matrix structure.}
    \label{fig:rsm_heatmaps}
\end{figure*}

Figure~\ref{fig:rsm_heatmaps} shows the target and decoded
representational similarity matrices for Subject~01.
Table~\ref{tab:supp-semantic-summary} summarizes the
four-subject 2AFC, CCD-H, and RSA results.

\begin{table}[!tb]
\centering
\scriptsize
\setlength{\tabcolsep}{2.5pt}
\renewcommand{\arraystretch}{1.02}
\begin{tabular*}{\columnwidth}{
@{\extracolsep{\fill}}
lrrr
@{}}
\toprule
\textbf{Metric}
& \textbf{MindEye2 (\%)}
& \textbf{Ours (\%)}
& \textbf{$\Delta$ (pp)} \\
\midrule
2AFC I$\rightarrow$B
& $90.68\pm6.40$
& \textbf{$96.78\pm0.99$}
& $+6.10\pm5.43$ \\
2AFC B$\rightarrow$I
& \textbf{$98.40\pm0.83$}
& $98.38\pm0.33$
& $-0.03\pm0.69$ \\
CCD-H
& $62.30\pm3.84$
& \textbf{$63.65\pm1.62$}
& $+1.35\pm2.98$ \\
RSA Pearson
& $28.05\pm2.40$
& \textbf{$35.50\pm1.30$}
& $+7.46\pm1.91$ \\
\bottomrule
\end{tabular*}
\caption{Four-subject mean $\pm$ standard deviation on the
982 shared test images. Values are reported in percent, and
differences are in percentage points.}
\label{tab:supp-semantic-summary}
\end{table}

\paragraph{Cross-subject IS-RSA.}
All six pairwise IS-RSA point differences are positive, as
shown in Table~\ref{tab:supp-isrsa-matrix}.

\begin{table}[!tb]
\centering
\footnotesize
\setlength{\tabcolsep}{4pt}
\begin{tabular*}{\columnwidth}{
@{\extracolsep{\fill}}
ccccc
@{}}
\toprule
\textbf{Subj / $\Delta$ (pp)}
& \textbf{1}
& \textbf{2}
& \textbf{5}
& \textbf{7} \\
\midrule
1 & -- & +6.50 & +7.63 & +8.88 \\
2 & +6.50 & -- & +6.10 & +6.87 \\
5 & +7.63 & +6.10 & -- & +5.67 \\
7 & +8.88 & +6.87 & +5.67 & -- \\
\bottomrule
\end{tabular*}
\caption{Pairwise IS-RSA differences in percentage points
($\Delta=\text{Ours}-\text{MindEye2}$ after multiplying
correlations by 100).}
\label{tab:supp-isrsa-matrix}
\end{table}

\subsection{Limited-Data Decoding}

The main paper reports CCD and 2AFC-Hard results for
Subjects~01 and~05 using one, two, and forty training
sessions. The differences are most visible in the one- and
two-session settings. Figure~\ref{fig:eff_2afc} gives an
enlarged view of the 2AFC-Hard results.

\par\medskip
\noindent
\begin{minipage}{\columnwidth}
\centering
\includegraphics[width=\linewidth]
{photos/Fig_efficiency_twoafc_hard_v2.png}
\captionof{figure}{Enlarged 2AFC-Hard data-efficiency results
for Subjects~01 and~05 using one, two, and forty training
sessions.}
\label{fig:eff_2afc}
\end{minipage}
\par\medskip

\subsection{Qualitative Example Selection}

The qualitative comparison in the main paper contains paired
Ground Truth, MindEye2, and ConceptAlign reconstructions. The
examples illustrate the semantic categories stated in the
figure: fine-grained details when both methods recover the
main scene, and differences in object identity, count,
arrangement, or visible attributes. They are not used to
estimate the frequency of these differences over the full
test set.

\section{Margin Ablation}
\label{supp:margin}

The margin $m$ controls the separation required between the
positive and counterfactual captions.
Table~\ref{tab:margin_ablation} reports the four tested
values.

\par\vspace{4pt}
\noindent
\begin{minipage}{\columnwidth}
\centering
\footnotesize
\setlength{\tabcolsep}{4pt}
\renewcommand{\arraystretch}{1.05}

\begin{tabular*}{\linewidth}{
@{\extracolsep{\fill}}
lcccc
@{}
}
\toprule
\textbf{Margin ($m$)}
& .05
& \textbf{.10}
& .20
& .50 \\
\midrule
CCD@1 (\%)
& 59.20
& \textbf{63.60}
& 62.10
& 61.00 \\
Retrieval FWD@1 (\%)
& 17.00
& \textbf{18.00}
& \textbf{18.00}
& 16.00 \\
\bottomrule
\end{tabular*}

\captionof{table}{Sensitivity to the ranking margin $m$.
Values are reported in percent.}
\label{tab:margin_ablation}
\end{minipage}

\par\vspace{8pt}

\section{Experimental Settings}
\label{supp:repro}
\enlargethispage{3\baselineskip}

\subsection{Software and Hardware}

Backbone training, ConceptAlign fine-tuning, and
reconstruction inference were run on a single NVIDIA
A100-SXM4 GPU with 80\,GB of memory. The environment used
Ubuntu 22, NVIDIA driver 550.127.05, CUDA 12.4, Python 3.12.7,
PyTorch 2.5.0, NumPy 2.1.3, and cuDNN 9.1.0.70.
OpenCLIP-based CCD evaluation was run on a single NVIDIA RTX
3090 GPU with 24\,GB of memory using
\texttt{open\_clip} 3.3.0 and the ViT-bigG-14 encoder. Other
metrics and statistical analyses were computed from exported
predictions and features.

\subsection{Common Training Settings}

The controlled variants use the same Subject-01 data,
backbone initialization, training budget, and evaluation
pipeline. The controlled fine-tuning set contains 8,946 fMRI
samples, of which 909 captioned entries provide
counterfactual text supervision. All samples contribute to
the inherited reconstruction objectives.
Table~\ref{tab:config_summary} reports the main settings.

\par\medskip
\noindent
\begin{minipage}{\columnwidth}
\centering
\scriptsize
\setlength{\tabcolsep}{2.6pt}
\renewcommand{\arraystretch}{0.96}
\begin{tabular*}{\linewidth}{
@{\extracolsep{\fill}}
L{0.58\linewidth}
R{0.32\linewidth}
@{}}
\toprule
\textbf{Item} & \textbf{Setting} \\
\midrule
Training subject / sessions & S01 / 40 \\
fMRI training samples & 8,946 \\
Caption-supervised entries & 909 \\
Batch size & 32 \\
Optimizer & AdamW \\
Weight decay & .01 \\
Gradient-norm clip & 1.0 \\
Initialization epochs / max LR & 30 / $3\times10^{-4}$ \\
Fine-tuning epochs / max LR & 120 / $1\times10^{-4}$ \\
Scheduler & LinearLR \\
Training seed & 42 \\
Precision & BF16 \\
Text alignment weight & .05 \\
Text temperature $\tau$ & .07 \\
Hard-negative scale & .30 \\
Ranking margin $m$ & .10 \\
Training negatives & Top-1 \\
\bottomrule
\end{tabular*}
\captionof{table}{Main settings for the controlled
negative-source study. C1 removes hard-negative supervision;
C2, C3, and ConceptAlign differ in the negative source.}
\label{tab:config_summary}
\end{minipage}
\par\medskip

The inherited loss weights are 1 for the CLIP branch, 30 for
the diffusion-prior branch, and .5 for the blurry
reconstruction branch. Bias and LayerNorm parameters use zero
weight decay. Optimization is performed every iteration, and
mixup is active during the first third of fine-tuning.

\vspace{-0.25em}
\subsection{Inference Settings}
\vspace{-0.15em}

For each unique test stimulus, its three fMRI repetitions are
processed independently through the subject-specific ridge
mapping and the shared visual-token backbone. The resulting
outputs are averaged before reconstruction. The averaged
backbone tokens condition a diffusion prior sampled for 20
steps with condition scale 1, producing
$256\times1664$ predicted visual tokens.

The predicted tokens condition an SDXL-derived unCLIP decoder,
which starts from random latent noise and generates one
initial reconstruction using 38 Euler--EDM sampling steps with
classifier-free guidance 5.0. The same tokens are also passed
to the auxiliary GIT caption decoder inherited from MindEye2.
The predicted caption can be saved as an auxiliary output and,
following the MindEye2 protocol~\cite{scotti2024mindeye2},
can optionally be used with the initial reconstruction in an
SDXL image-to-image refinement stage. This optional
post-processing does not change the decoded visual tokens or
the representation-level evaluation.

ConceptAlign adds no counterfactual-caption generation or
external language-model call at inference. Inference uses
seed 42 and generates one unCLIP sample per stimulus before
any optional post-processing.

\vspace{-0.35em}
\subsection{Statistical Evaluation}
\vspace{-0.2em}

Paired comparisons use 10,000 image-level bootstrap resamples
with seed 2027 and report 95\% percentile confidence
intervals. The same resampled image indices are used for both
models in each comparison, with
$\Delta=\text{Ours}-\text{Control}$.

\end{document}